\documentclass[sigconf]{acmart}

\usepackage{graphicx}
\usepackage{multicol} 
\usepackage{balance}  

\usepackage{caption}
\usepackage{subcaption}
\usepackage{soul}
\usepackage[linesnumbered,ruled]{algorithm2e}
\usepackage{ifthen}
\usepackage{amsmath}
\usepackage{amsfonts}
\usepackage{booktabs}
\usepackage{algorithmic}
\SetKwInput{KwInput}{Input}                
\SetKwInput{KwOutput}{Output}              

\usepackage{booktabs} 
\usepackage{enumitem}
\usepackage{comment}
\usepackage{xcolor}
\usepackage[utf8]{inputenc}

\usepackage{hyperref}

\usepackage{color, colortbl}
\definecolor{Gray}{gray}{0.9}

\usepackage{multirow}

\usepackage{xcolor,colortbl}

\definecolor{Gray}{gray}{0.70}
\definecolor{Gray2}{gray}{0.90}
\definecolor{LightCyan}{rgb}{0.88,1,1}

\theoremstyle{definition}
\newtheorem{definition}{Definition}[section]

\usepackage{relsize}

\usepackage{setspace}
\usepackage{microtype}

\newcolumntype{b}{>{\columncolor{Gray}}c}
\newcolumntype{a}{>{\columncolor{Gray2}}c}
\newcolumntype{d}{>{\columncolor{LightCyan}}c}

\copyrightyear{2022}
\acmYear{2022}
\setcopyright{rightsretained}
\acmConference[SIGSPATIAL '22]{The 30th International Conference on Advances in Geographic Information Systems}{November 1--4, 2022}{Seattle, WA, USA}
\acmBooktitle{The 30th International Conference on Advances in Geographic Information Systems (SIGSPATIAL '22), November 1--4, 2022, Seattle, WA, USA}
\acmDOI{10.1145/3557915.3560943}
\acmISBN{978-1-4503-9529-8/22/11}

\begin{document}
\setlength{\pdfpagewidth}{8.5in}
\setlength{\pdfpageheight}{11in}

\title[Will there be a construction?]{Will there be a construction? Predicting road constructions based on heterogeneous spatiotemporal data}

\author[ ]{Amin Karimi Monsefi}
\affiliation{%
  \institution{The Ohio State University}
  \streetaddress{2015 Neil Ave}
  \city{Columbus}
  \state{Ohio}
  \postcode{43210}
}
\email{karimimonsefi.1@osu.edu}

\author[ ]{Sobhan Moosavi}
\affiliation{%
  \institution{Lyft Inc.}
  \streetaddress{185 Berry Street}
  \city{San Francisco}
  \state{California}
  \postcode{94107}
}
\email{smoosavi@lyft.com}

\author[ ]{Rajiv Ramnath}
\affiliation{%
  \institution{The Ohio State University}
  \streetaddress{2015 Neil Ave}
  \city{Columbus}
  \state{Ohio}
  \postcode{43210}
}
\email{ramnath.6@osu.edu}

\begin{abstract}
Road construction projects maintain transportation infrastructures. These projects range from the short-term (e.g., resurfacing or fixing potholes) to the long-term  (e.g., adding a shoulder or building a bridge). Deciding what the next construction project is and when it is to be scheduled is traditionally done through inspection by humans using special equipment. This approach is costly and difficult to scale. An alternative is the use of computational approaches that integrate and analyze multiple types of past and present spatiotemporal data to predict location and time of future road constructions. This paper reports on such an approach, one that uses a deep-neural-network-based model to predict future constructions. Our model applies both convolutional and recurrent components on a heterogeneous dataset consisting of construction, weather, map and road-network data. We also report on how we addressed the lack of adequate publicly available data - by building a large scale dataset named ``US-Constructions'', that includes 6.2 million cases of road constructions augmented by a variety of spatiotemporal attributes and road-network features, collected in the contiguous United States (US) between 2016 and 2021.  Using extensive experiments on several major cities in the US, we show the applicability of our work in accurately predicting future constructions - an average f1-score of $0.85$ and accuracy $82.2\%$ - that outperform baselines. Additionally, we show how our training pipeline addresses spatial sparsity of data.
\end{abstract}


\maketitle

\section{Introduction}
\label{sec:intro}
Road constructions are essential to transportation infrastructures. Recently released data by the United States Census Bureau showed that the annual value of road constructions increased from 87.9 billion dollars in 2017 to 100.4 billion dollars in 2021 - an over 18\% increase in just five years\footnote{Visit \url{https://www.census.gov/construction/c30/prpdf.html} for detailed reports.}. In order to make the best use of the substantial amount of capital invested in this sector, it is crucial to determine construction sites wisely. Deciding what the next construction project is and when it
is to be scheduled is traditionally done through inspection by humans using special equipment - an approach that is  expensive and limited in coverage. We therefore explored an alternative for this paper, namely, the use of computational solutions to determine future constructions. 

To our knowledge, determining when and where road constructions are needed via computational solutions is a relatively new, less explored research area. Existing studies have focused on image analysis, such as for detecting cracks and potholes in road surface \citeN{guan2021automated, bibi2021edge, arya2021rdd2020, wen2022automated} and detecting road closures via telematics and vehicle probe data \citeN{kataoka2018smartphone}. While promising, the lack of extensive input data (e.g., fine-grained satellite or street imagery at scale) has limited the applicability and extensibility of these approaches. Additionally, the data used are mostly private, which limits both independent validation and the ability to build on and extend prior work. In general, current computational solutions have been limited due to the need for data to build good models with real-world applicability.

The work presented here begins by addressing challenges with data. To this end, this paper introduces a unique dataset of 6.2 million road constructions in the United States between 2016 and 2021. Our dataset offers a variety of contextual details for each construction, including location, time, a brief human provided description, daylight and weather at the start of each construction, and several map related features that contextualize the location of a construction (e.g., if it is close to a highway junction, road-type, etc.). We also carefully describe our process for building this dataset. Thus, researchers may either directly use our dataset or mimic our approach and build their own dataset.

Next, we explore an important and useful research problem, that of ``identifying future constructions from past constructions along with their spatiotemporal context (e.g., traffic, weather, and map imagery) for certain locations (represented by their geographical region hexagon -- see Section~\ref{sec:problem}) during specific time frames (e.g., the next 15 days).  Our goal is to develop a cost-saving, coverage-enhancing approach complementary to current practices. For example, we see our approach as being used to quickly identify potential sites, which can then be evaluated by human inspections.

In our approach, we model heterogeneous spatiotemporal information using a deep-neural-network that combines \textit{recurrent} and \textit{convolutional} components. The convolutional component is used for extracting latent information from map imagery. The recurrent component models time-series data (e.g., traffic and weather) along with additional spatial information (e.g., features of the road-network) about a location. The output of these components are then concatenated and fed to a fully connected component which produces the final output. Our goal with this model is to predict the possibility of a construction event in the near term (specifically, the next 15 days). While this formulation best suits short-term constructions (i.e., those that take a few hours to a few days), we believe it is very useful effective, given that the majority of constructions in our data can be considered as short-term (see Section \ref{sec:dataset}). Extensive experiments and results demonstrate improvements over traditional and neural-network-based machine learning baselines. On average, our proposed model outperformed the best baseline model by $3.2\%$ in accuracy and by $2.8\%$ in F1-score, when tested over multiple major cities in the United States. Our model also demonstrated robustness in dealing with \textit{spatial sparsity} in tests at the state level - a real-world scenario of training data being only available for parts of a region. 
In summary, the main contributions of this paper are:
\begin{itemize}
    \item \textbf{Dataset}: We introduce a new dataset of road constructions and closures for the continental United States, with about $6.2$ million cases from the years $2016$ to $2021$. To our knowledge, this is the first public dataset that offers this type and scale of data. 
    \item \textbf{Insights}: We glean a variety of insights by analyzing the US-Constructions dataset. We detail these insights with a view to inspiring  other researchers to use our data for other applications, especially those aimed at enhancing transportation infrastructures and their safety. 
    \item \textbf{Model}: We present a deep neural network model to predict short-term constructions. Our model is capable of using heterogeneous data, and resulted in superior prediction outcomes when compared to several state-of-the-art traditional and deep-learning models. 
\end{itemize} 
\section{Related Work}
\label{sec:related}

We examined several previous studies on road construction issues. These studies examined a variety of topics ranging from detection of road issues (such as cracks) \citeN{han2021change}, analysis of the maintenance of roads \citeN{issa2017evaluation, sunitha2013application, li2019applications, shtayat2020review, krishna2022sustainable}, prediction and management of the costs of road construction \citeN{moretti2016management, chong2016international, petroutsatou2012early, onyango2018analysis, nahvi2021data}, and lifecycle analysis of roads \citeN{hoxha2021life, gulotta2019life, jiang2019estimation, hasan2019critical, trunzo2019life}. We highlight some of this work next.

Tong et al. \citeN{tong2018recognition} employed deep convolutional neural network (DCNN) models for finding the length of cracks in asphalt pavement from gray-scale images. This work used a dataset of $8,000$ images, $7,500$ images with cracks and $500$ images without. They classified images to 8 different classes according to the length of the crack in centimeters, achieving an accuracy of $94.36\%$ in this classification. 
In another study, Ye et al. \cite{ye2019convolutional} employed a convolutional neural network (CNN) model to identify potholes in asphalt pavements. They used a dataset of 400 images that were collected from different pavements under different lighting conditions. These images were cropped into $96,000$ smaller images of size $256 \times 256$ pixels. Authors reported $98.95\%$ precision to detect potholes, and their stability study suggested robustness in various real-world settings. 

Automatic detection of road-closures is the topic of another group of studies \citeN{pietrobon2019algorithm, cheng2017automatic}. 
Cheng et al. \citeN{cheng2017automatic} presented a high-efficiency road closure detection framework based on multi-feature fusion. Their framework had two parts, an offline road closure feature modeling part and an online identification part. For the offline modeling, they first partitioned the road-network into grids, and then extracted their road closure features of these grids and the roads within them from historical data. The online component screened out closed grid candidates based on the plunge in traffic flow. They also identified sections with road closures based on variations in turning behavior by drivers on these roads. Their framework was evaluated on three real-world datasets - from Chengdu, Shanghai, and Beijing.

With respect to the life-cycle analysis (LCA) of roads  Gulotta et al. \citeN{gulotta2019life} applied a life-cycle approach to evaluate the energy and environmental impacts of a typical urban road in Italy. They evaluated the energy and environmental impact of various bituminous combinations. For each analyzed scenario, the contribution of each life-cycle phase to the total effects and to the energy and environmental hot-spots were identified as opportunities for improvement. 
Jiang et al. \citeN{jiang2019estimation} reviewed and analyzed 94 papers that adopted LCA methods to assess the environmental effects over the life cycle of roads. Their study resulted in multiple outcomes including identifying limitations and challenges of using LCA in the environmental assessment of roads, as well as in identifying future research directions.

Our paper borrows concepts from the papers described above while tackling the different problem of predicting the possibility of a future construction event at a location. To our knowledge, this is the first study that seeks to solve a problem of this type using a purely computational approach. This work has real-world application, in that it can be used to find areas in need of maintenance. It is also cost-effective because it is based on analyzing already recorded and available information such as past road constructions, road-network features (e.g., road class data, average speed, and map annotations), weather data, and coarse-grained map images. As we show in this paper, the input data we employ is easy to collect and available to the public. This is in contrast to those studies so far, which utilized private or extensive datasets. 

\section{The US-Constructions Dataset}
\label{sec:dataset}
\begin{figure}[ht!]
    \centering
    \includegraphics[scale=0.4]{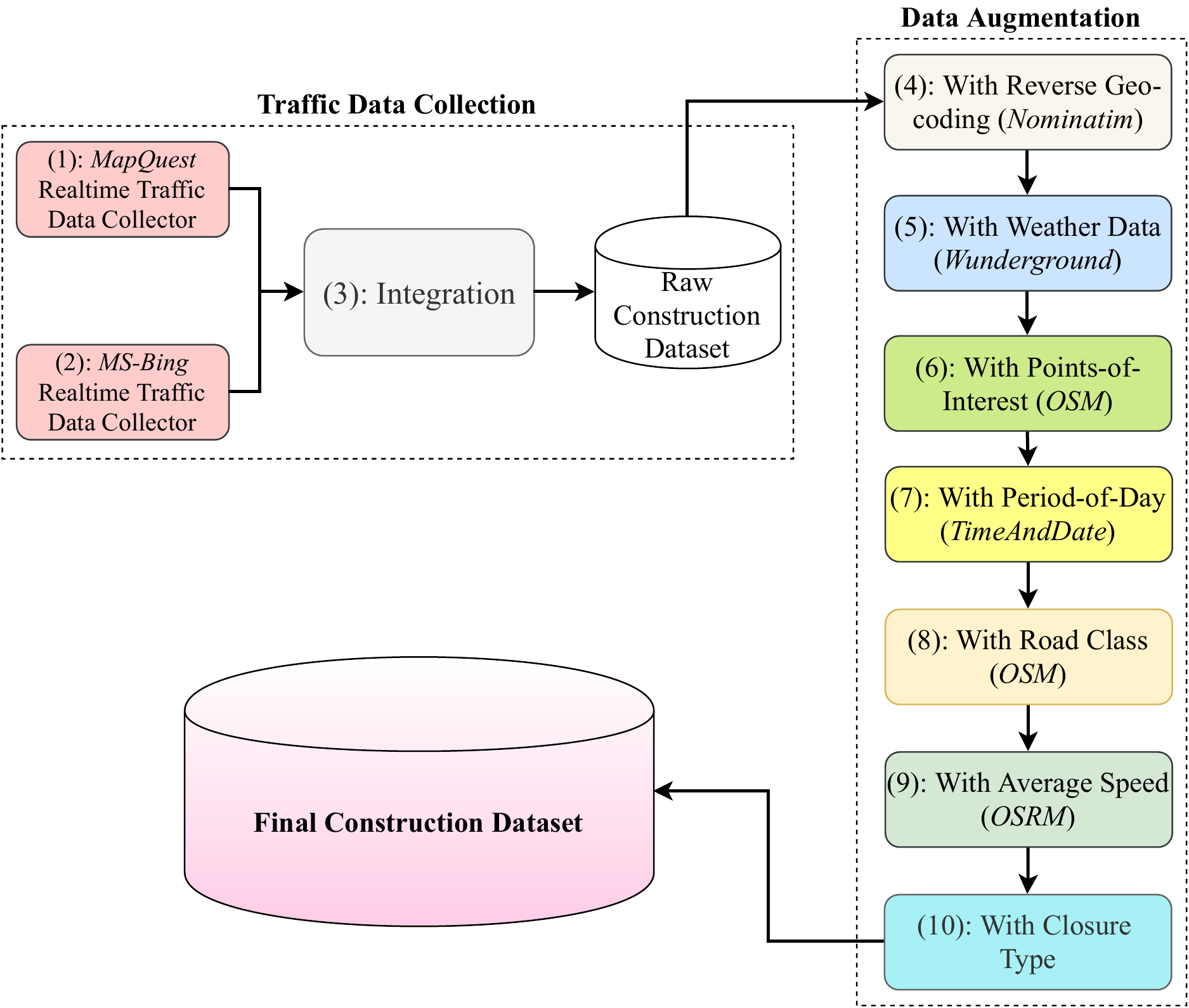}
    \caption{Process of Building US-Constructions Dataset}
    \label{fig:dataset_process}
\end{figure}

In this section, we describe the countrywide traffic construction dataset, which we have named \textit{US-Constructions}, and our process for building it. The ten major steps in this process are shown in Figure~\ref{fig:dataset_process} and described in this section. The resulting dataset contains 6.2 million road constructions that took place in the continental United States between January 2016 and December 2021, and is publicly available at \url{https://www.kaggle.com/datasets/sobhanmoosavi/us-road-construction-and-closures}. 

\subsection{Collecting Raw Construction Data}
We collected streaming traffic data from two real-time data providers - ``MapQuest Traffic'' \cite{mapquest} and ``Microsoft Bing Maps'' \cite{bing}. These providers broadcast real-time road construction and closure data collected by a variety of entities (e.g., law enforcement agencies, traffic cameras, and traffic sensors). We integrated the raw data to remove duplicates using a set of heuristic spatiotemporal filters. 0.63 million cases were pulled from MapQuest, and 5.54 million cases from BingMaps, with less than $1\%$ overlap between the two sources.

\subsection{Augmentation with Reverse Geocoding}
This is the process of translating raw location data (i.e., GPS coordinates) to addresses that included elements such as {\em street number}, {\em street name}, {\em city}, {\em state}, and {\em zip-code}. We employed the {\em Nominatim} tool \cite{nominatim} to perform this reverse geocoding. 

\subsection{Augmentation with Weather Data}
Weather information is a useful context for traffic constructions (in particular short-term cases). We employed {\em Weather Underground} API  \cite{wunderground} to obtain the weather during each construction. Raw weather data was collected from $2,072$ airport weather stations from around the United States. This raw data comes in the form of observation records, where each record consists of several attributes such as {\em temperature}, {\em humidity}, {\em wind-speed}, {\em pressure}, {\em precipitation} (in millimeters), and {\em condition}\footnote{Possible values are {\em clear}, {\em snow}, {\em rain}, {\em fog}, {\em hail}, and {\em thunderstorm}.}. We collected several records per day from each station. Note that each record is generated when any significant change occurs in any of the measured weather attributes. 

Each construction event $c$ was then augmented with weather data as follows. First the closest weather station $s$ was identified. Then, of the weather observation records reported from $s$, we looked for the weather observation record $w$ whose reported time was closest to the start time of $c$. $c$ was then augmented with $w$. 

\subsection{Augmentation with POI Annotation}
Points-of-interest (POI) are locations annotated on a map as {\em amenities}, {\em traffic signals}, {\em crossings}, etc. These annotations are associated with the {\em nodes} on a road network. A node can be associated with many POI tags; in this work, we used the 12 tags described in Table~\ref{tab:poi_types}. We obtained these tags from the OpenStreetMap (OSM) \cite{osm} system, and for the continental United States. The applicable POI annotations for a traffic construction $c$ are those that are located within a threshold distance $\tau$ from $c$. We adopted the same threshold as in \cite{moosavi2019accident}. 

\begin{table}[ht]
    \small 
    \setlength\tabcolsep{2pt}
    \centering
    \caption{Definition of Point-Of-Interest (POI) annotations based on OpenStreetMap (OSM).}\vspace{-5pt}
    \begin{tabular}{| c | c|}
        \rowcolor{Gray}
        \hline
        \textbf{Type} &  \textbf{Description}\\
        \hline
        Amenity & \begin{tabular}{@{}c@{}} Refers to particular places such as restaurant,\\ library, college, bar, etc.\end{tabular} \\ 
        \hline
        Bump & Refers to speed bump or hump to reduce the speed. \\ 
        \hline
        Crossing & \begin{tabular}{@{}c@{}} Refers to any crossing across roads for \\ pedestrians, cyclists, etc.\end{tabular} \\ 
        \hline
        Junction & Refers to any highway ramp, exit, or entrance. \\ 
        \hline
        No-exit & \begin{tabular}{@{}c@{}} Indicates there is no possibility to travel further\\ by any transport mode along a formal path or route.\end{tabular}  \\ 
        \hline
        Railway & Indicates the presence of railways. \\ 
        \hline
        Roundabout & Refers to a circular road junction.\\  
        \hline
        Station & Refers to public transportation stations (bus, metro, etc.). \\ 
        \hline
        Stop & Refers to stop signs \\ 
        \hline
        Traffic Calming & Refers to any means for slowing down traffic speed. \\ 
        \hline
        Traffic Signal & Refers to traffic signals on intersections \\ 
        \hline
        Turning Loop & \begin{tabular}{@{}c@{}} Indicates a widened area of a highway with\\a non-traversable island for turning around. \end{tabular}  \\ 
        \hline
    \end{tabular}
    \label{tab:poi_types}
\end{table}

\subsection{Augmentation with Period of Day}
Given the start time of a construction record, we used ``TimeAndDate'' API \cite{timeanddate} to label it as {\em day} or {\em night}. We assign this label based on four different daylight systems, namely {\em Sunrise/Sunset}, {\em Civil Twilight}, {\em Nautical Twilight}, and {\em Astronomical Twilight}. 

\subsection{Augmentation with Road Class}
\label{sec:dataset_road_class}
Road class (e.g., primary, secondary, and motorway) is an important feature of the location of a construction. We used OSM to obtain this information and adopted its road classification system\footnote{See \url{https://wiki.openstreetmap.org/wiki/Key:highway} for a list of OSM-based road types.} to annotate constructions. We describe our annotation process below. 

For a given construction $c$, the goal is to find the most relevant road class based on its start and end locations\footnote{If the end location is not available, then we only use the start location}. The OSM map data contains {\it nodes} and {\it ways}. A node is a single point defined by latitude, longitude, and a node id. A way represents a route in a road-network by an ordered list of nodes. We map location of a construction to an OSM node by using the ``Nearest Service'' OSRM APIs\footnote{See \url{http://project-osrm.org/docs/v5.5.1/api/\#nearest-service}} to find the $S$ nearest nodes to the start and end location of a construction (we empirically set $S=10$). After finding the $S$ nearest nodes, we prune outlier nodes using an aggressive distance threshold $D=50 \textit{meters}$. This results in a set $\mathcal{N}$ of nodes nearest to the location of a construction. Next, for each node $n \in \mathcal{N}$, we find a set $W_n$ of the OSM map "ways" that contain $n$, and build an aggregated set $\mathcal{W} = \bigcup_{n \in \mathcal{N}} W_n$. For each way $w \in \mathcal{W}$, we then use the OSM service ``Full`` to obtain road-class information\footnote{See \url{https://wiki.openstreetmap.org/wiki/API_v0.6} and check out \textit{GET /api/0.6/[way|relation]/\#id/full}}. Finally, we look for the most frequent road class that was returned for the ways in $\mathcal{W}$, and annotate the construction $c$ with that. 

\subsection{Augmentation with Average Road Speed}
Next, we obtain the average speed for each construction event from its start and end locations. For the about $10\%$ of constructions that do not have an end location, we do not infer an average speed. Instead, we use the OSM ``Route Service'' \footnote{see \url{http://project-osrm.org/docs/v5.5.1/api/\#route-service}} to estimate a free-flow speed for the roads with those constructions. 

\subsection{Augmentation with Closure Type}
Some of the constructions could result in \textit{road closures}. We introduce a rule-based process to annotate each construction with a closure type, if there is one. To begin with, we define three cases: {\it road-closure}, {\it lane-closure}, and {\it no-closure}. Our process of closure-based-annotation uses the human-provided description of each construction event, and employs several {\it regular-expression patterns} to infer the closure type. Examples of human-provided descriptions are listed in Table~\ref{tab:closure_examples}.

\begin{table}[h!]
    \small
    \centering
    \setlength\tabcolsep{2pt}
    \caption{Samples of human-provided description for some construction events that resulted in closure. ``Source'' shows which data provider was used to collect the data.} \vspace{-5pt}
    \begin{tabular}{|c|c|}
    \hline
        \rowcolor{Gray} \textbf{Description of Construction Event}&  \textbf{Source} \\
    \hline
        Closed due to roadwork & $Bing$ \\
    \hline
        Closed for bridge demolition work & $Bing$ \\
    \hline
        Right lane blocked. & $Bing$ \\
    \hline
        Two lanes blocked & $Bing$ \\
    \hline
        Roadway reduced to 1 lane & $Bing$ \\
    \hline
        Lane blocked. One lane closed & $Bing$ \\
    \hline
        Lane closed due to construction work & $MapQuest$ \\
    \hline
        Shoulder closed on entry ramp due to construction work & $MapQuest$ \\
    \hline
        Intermittent lane closures due to utility work & $MapQuest$ \\
    \hline
        Three lanes closed due to construction work & $MapQuest$ \\
    \hline
        Right lane closed due to construction work  & $MapQuest$ \\
    \hline
        One lane closed due to maintenance work & $MapQuest$ \\
    \hline
    \end{tabular}
    \label{tab:closure_examples}
    \vspace{-5pt}
\end{table}

From manual probing of the human-provided descriptions for a large set of ~10,000 randomly selected construction events we found 14 distinct patterns that represent closures. Of these patterns, four represent a {\it road closure} and ten of them a {\it lane closure}. Table~\ref{tab:closure_pattern} shows these patterns along with corresponding data source, as well as closure-type. 

\begin{table}[ht!]
    \centering
    \setlength\tabcolsep{10pt}
    \caption{Regular expression patterns to annotate closure type for construction events} \vspace{-5pt}
    \begin{tabular}{|c|c|c|}
    \hline
    \rowcolor{red!20} \textbf{Pattern}&  \textbf{Source} & \textbf{Closure Type} \\
    \hline
        close* * roadwork &  $Bing$ & $Road$ \\
    \hline
        close* * bridge  &  $Bing$ & $Road$ \\
    \hline
        hard shoulder close* &  $Bing$ & $Lane$ \\
    \hline
        * lane* block* &  $Bing$ & $Lane$ \\
    \hline
        * reduced * lane* &  $Bing$ & $Lane$ \\
    \hline
        * lane* close* &  $Bing$ & $Lane$ \\
    \hline
        close* * roadwork & $MapQuest$ & $Road$ \\
    \hline
        road close* * & $MapQuest$ &  $Road$ \\
    \hline
       * lane closure*  & $MapQuest$ & $Lane$ \\
    \hline
        hard shoulder block* & $MapQuest$ &  $Lane$\\
    \hline
        * reduced * lane* & $MapQuest$ & $Lane$ \\
    \hline
        * lane* block* & $MapQuest$ &  $Lane$\\
    \hline
       * lane* close*  & $MapQuest$ & $Lane$ \\
    \hline
       * shoulder close*  & $MapQuest$ &  $Lane$\\
    \hline
    \end{tabular}
    \label{tab:closure_pattern}
    \vspace{-5pt}
\end{table}

To evaluate the effectiveness of our patterns in finding closure types, we randomly selected $1,000$ construction events and annotated them with the derived patterns, which resulted in 43 road closures and 137 lane closures. After manually checking all resulting annotations and also cases without any annotations, we observed $precision = 100\%$ and $recall = 100\%$, which showed that our set of regular expression patterns was comprehensive and accurate. 

\subsection{Final Dataset}
The final dataset comprises 6.2 million construction records, collected between January 2016 and December 2021. Each construction record is described by 45 attributes as shown in Table~\ref{tab:final_dataset}. 

\begin{table}[ht!]
    \small
    \setlength\tabcolsep{.7pt}
    \centering
    \caption{US-Constructions Dataset (details as of Dec 2021)} \vspace{-5pt}
    \begin{tabular}{|>{\columncolor[gray]{0.9}}c|c|}
        \hline
        \cellcolor{blue!20}\textbf{Total Attributes} &  \cellcolor{blue!20} \textbf{45}\\
        \hline
        Traffic Attributes (8) &  \begin{tabular}{@{}c@{}} id, severity, start\_time, end\_time, \\ start\_point, end\_point, distance, and description \end{tabular} \\
        \hline
        Address Attributes (8) & \begin{tabular}{@{}c@{}} number, street, side (left/right), city, \\ county, state, zip-code, country  \end{tabular}\\
        \hline
        Weather Attributes (10) & \begin{tabular}{@{}c@{}} timestamp, temperature, wind\_chill, humidity,\\ pressure, visibility, wind\_direction, wind\_speed,\\precipitation, and condition (e.g., rain, snow, etc.) \end{tabular}\\
        \hline
        POI Attributes (12) & All cases in Table~\ref{tab:poi_types}\\
        \hline
        Period-of-Day (4) & \begin{tabular}{@{}c@{}} Sunrise/Sunset, Civil Twilight, \\Nautical Twilight, and Astronomical Twilight\end{tabular}\\
        \hline
        Other Attributes (3) & \begin{tabular}{@{}c@{}} Road class (e.g., highway, primary, etc.) \\ Average speed (e.g., 55 mph), and Closure type\end{tabular}\\
        \hline
        \hline
        \cellcolor{blue!20} \textbf{Total Constructions} &  \cellcolor{blue!20} \textbf{6.2 million}\\
        \hline
        \# MapQuest Constructions & 634k (10\%) \\
        \hline
        \# Bing Constructions & 5.54 million (90\%) \\
        \hline
        \# Reported by Both & ~ 1\% \\
        \hline
    \end{tabular}
    \label{tab:final_dataset}
\end{table}

We ran a meta-analysis on this dataset. Our findings from this analysis are detailed below: 
\begin{itemize}[leftmargin=*]
    \item \textbf{Top states and cities}: Figure~\ref{fig:top_states_cities} shows top states and cities, as well as top states when the number of constructions are normalized by the aggregate length of road network (in miles) for each state (data is taken from Federal Highway Administration site\footnote{see \url{https://www.fhwa.dot.gov/policyinformation/statistics/2020/hm60.cfm}}). Note how the state distribution changes after the normalization. 
    
        \begin{figure*}[h]
            \minipage{0.32\textwidth}
                \centering
                \includegraphics[width=\linewidth]{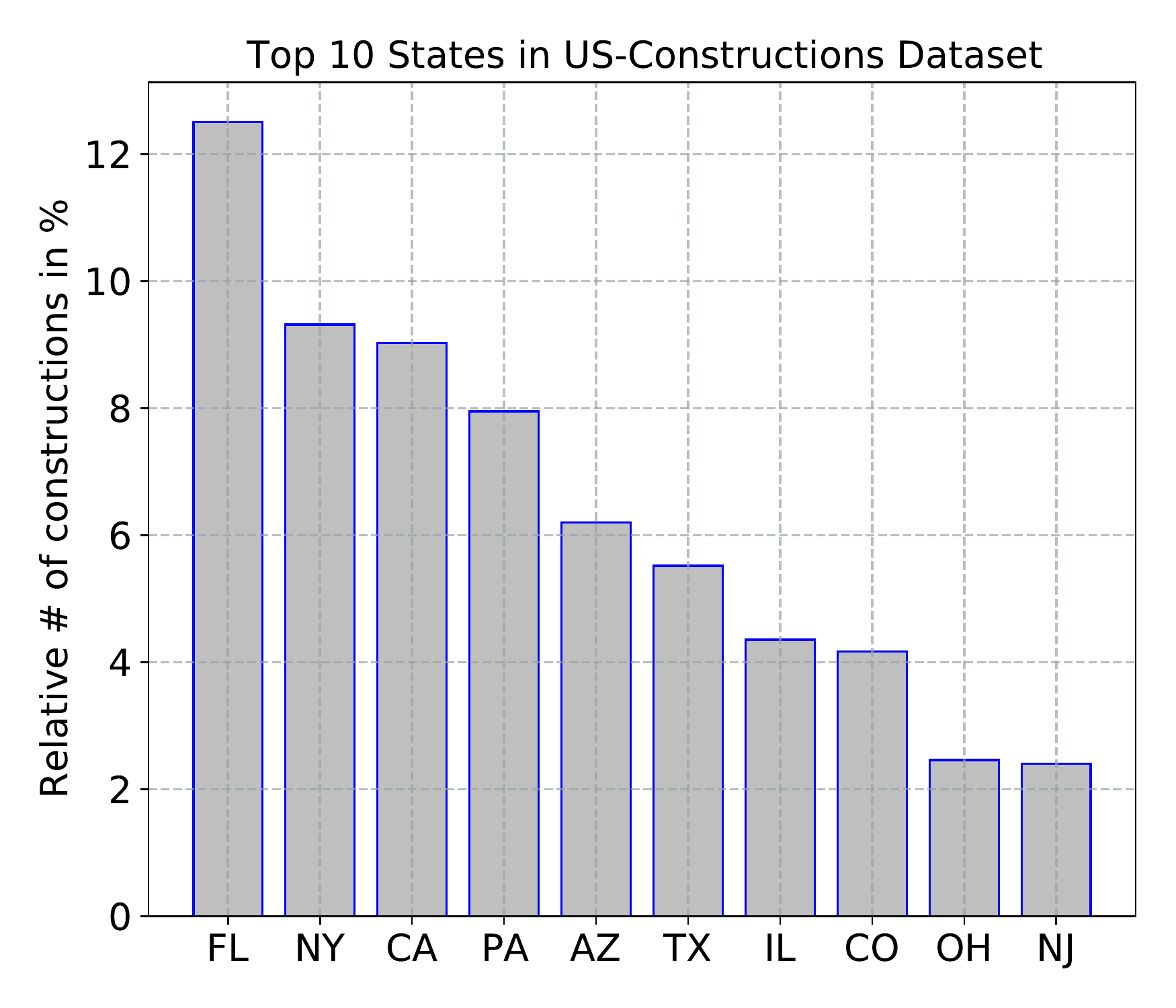}
                \small (a) Top States based on frequency
            \endminipage\hspace{5pt}
            \minipage{0.32\textwidth}
                \centering
                \includegraphics[width=\linewidth]{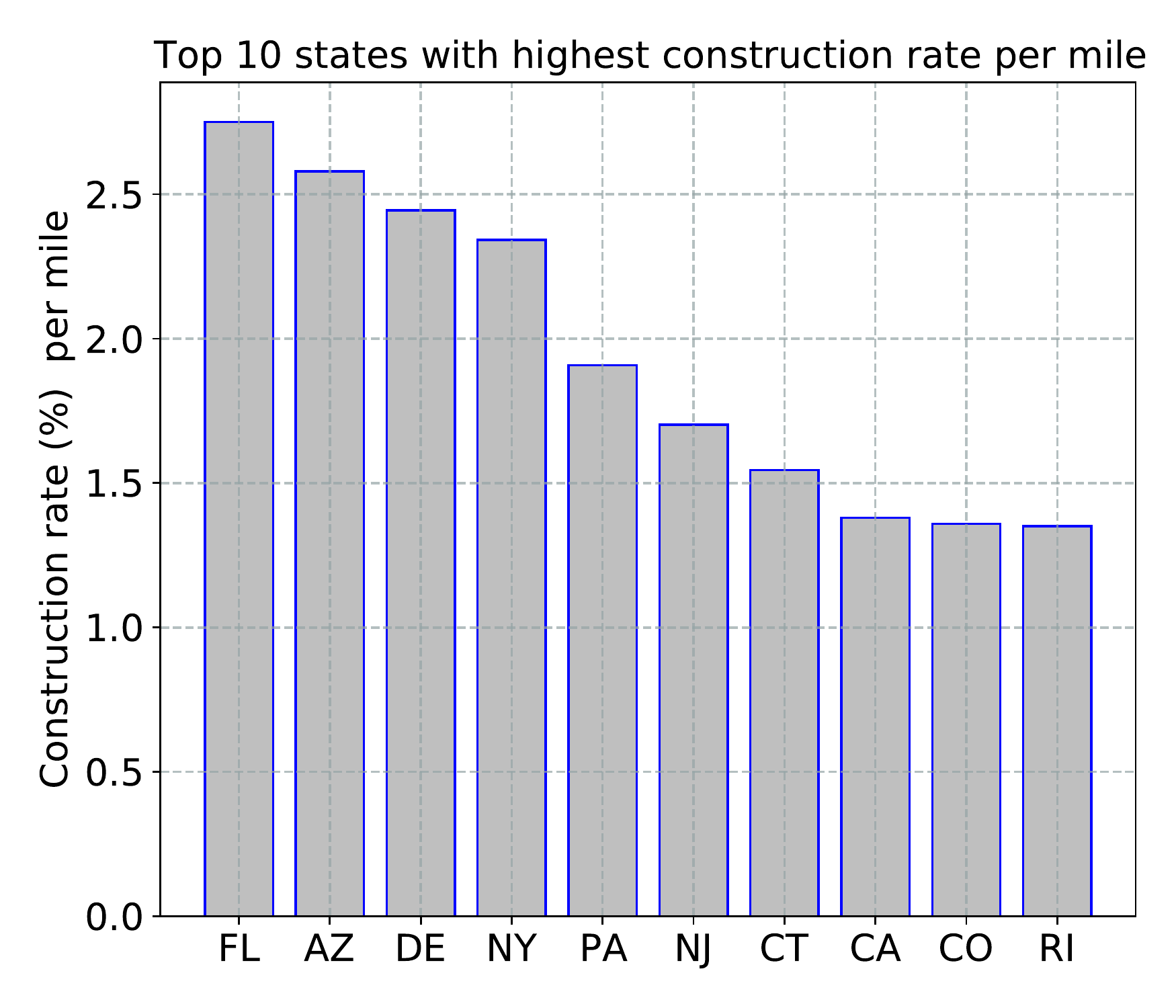}
                \small (b) Top states based on construction rate per mile
            \endminipage\hspace{5pt}
            \minipage{0.32\textwidth}
                \centering
                \includegraphics[width=\linewidth]{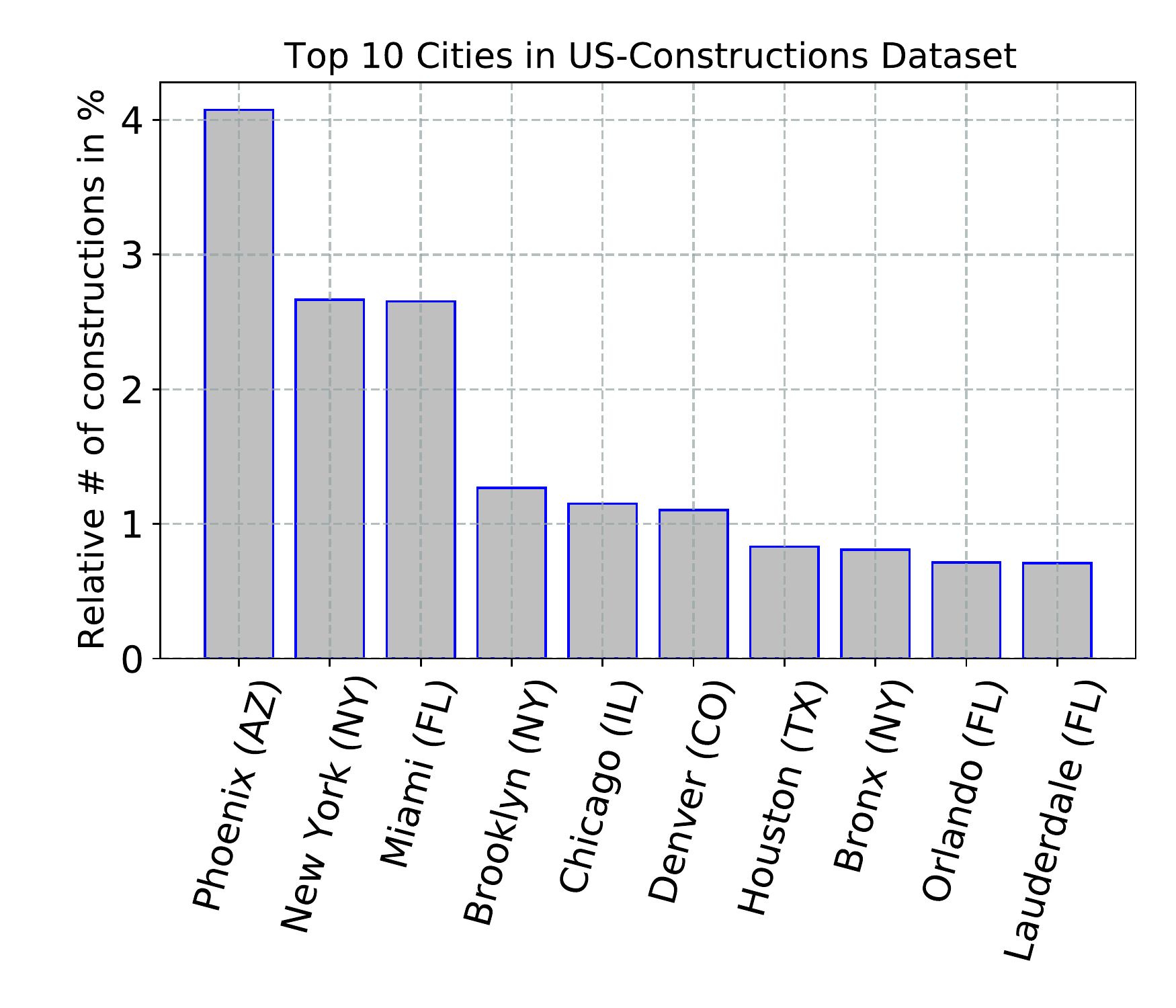}
                \small (c) Top Cities based on frequency
            \endminipage\hfill
            \vspace{-5pt}
            \caption{\small Construction distribution over top states (a), top states normalized by length of road network in miles (b), and top cities (c).}
            \label{fig:top_states_cities}
        \end{figure*}

    \item \textbf{Monthly distribution}: Figure~\ref{fig:other_analysis}(a) shows that we should expect to see more constructions start as well as finish towards the end of a year. The last four months of a year typically see more constructions finish than start, likely because of a pressure to finish ongoing work before the calendar year ends. 
    
        \begin{figure*}[h]
            \minipage{0.32\textwidth}
                \centering
                \includegraphics[width=\linewidth]{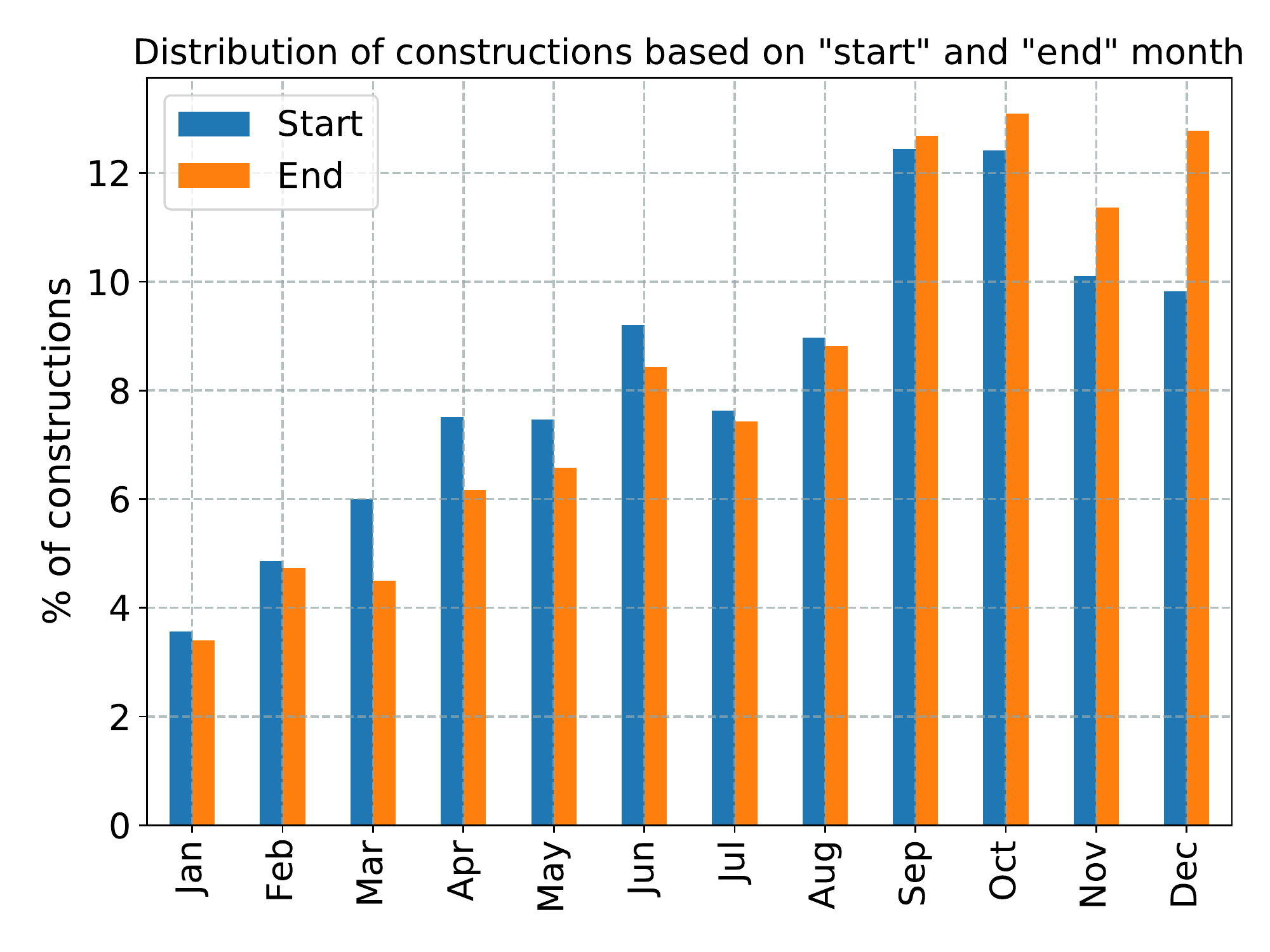}
                \small (a) Monthly distribution 
            \endminipage\hspace{5pt}
            \minipage{0.32\textwidth}
                \centering
                \includegraphics[width=\linewidth]{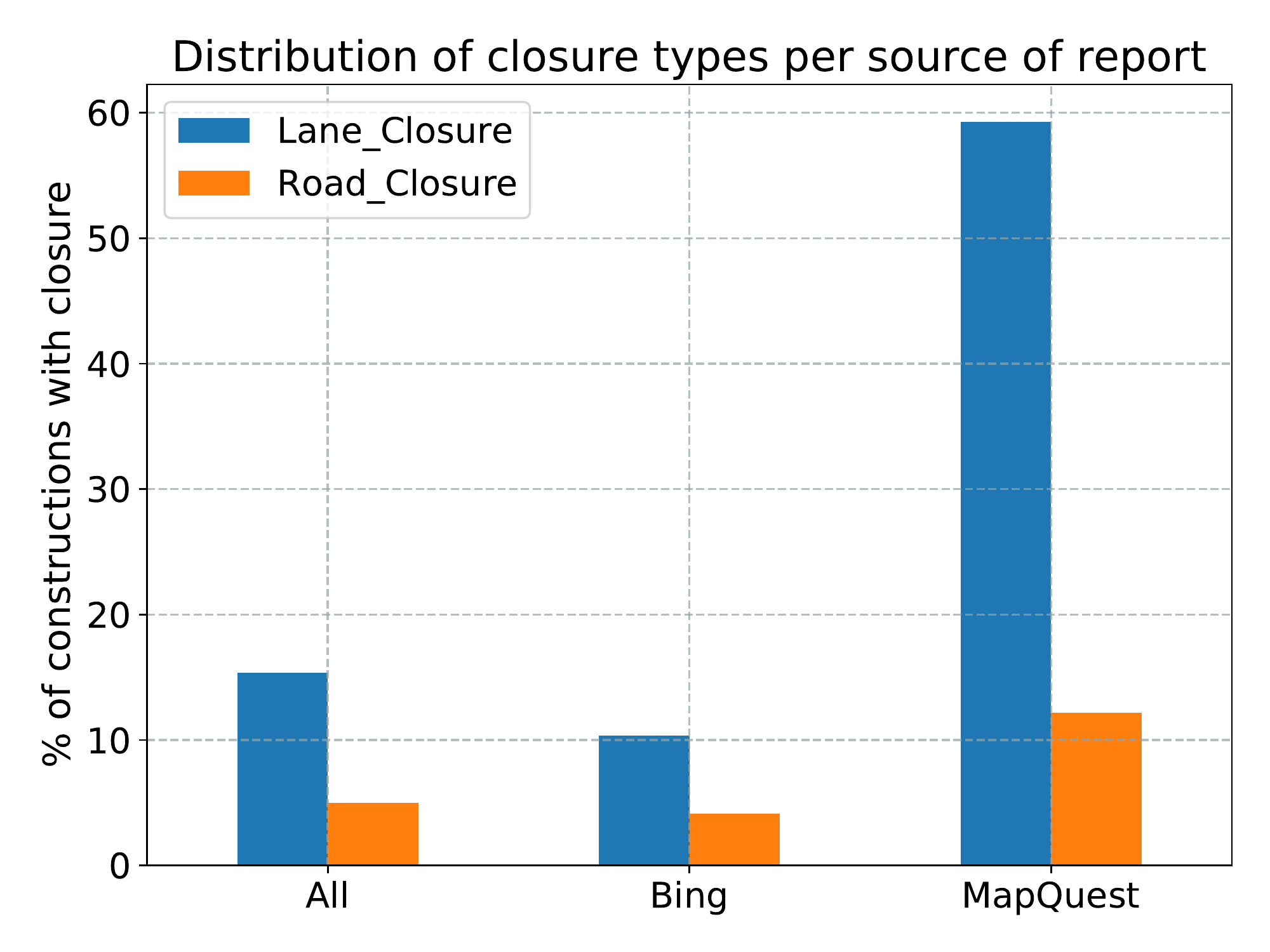}
                \small (b) Closure distribution
            \endminipage\hspace{5pt}
            \minipage{0.32\textwidth}
                \centering
                \includegraphics[width=\linewidth]{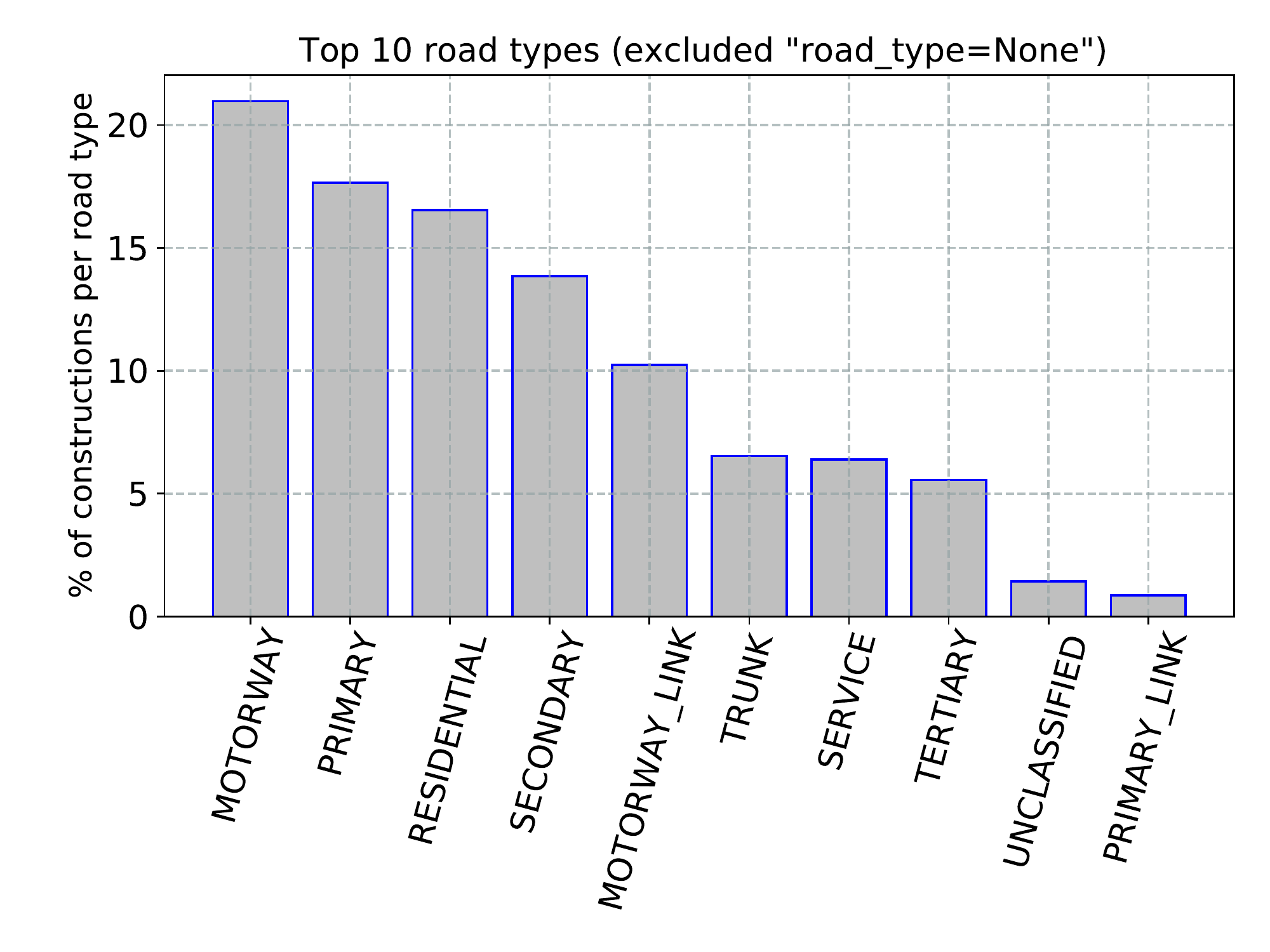}
                \small (c) OSM-based road type distribution
            \endminipage\hfill
            \minipage{0.32\textwidth}
                \centering
                \includegraphics[width=\linewidth]{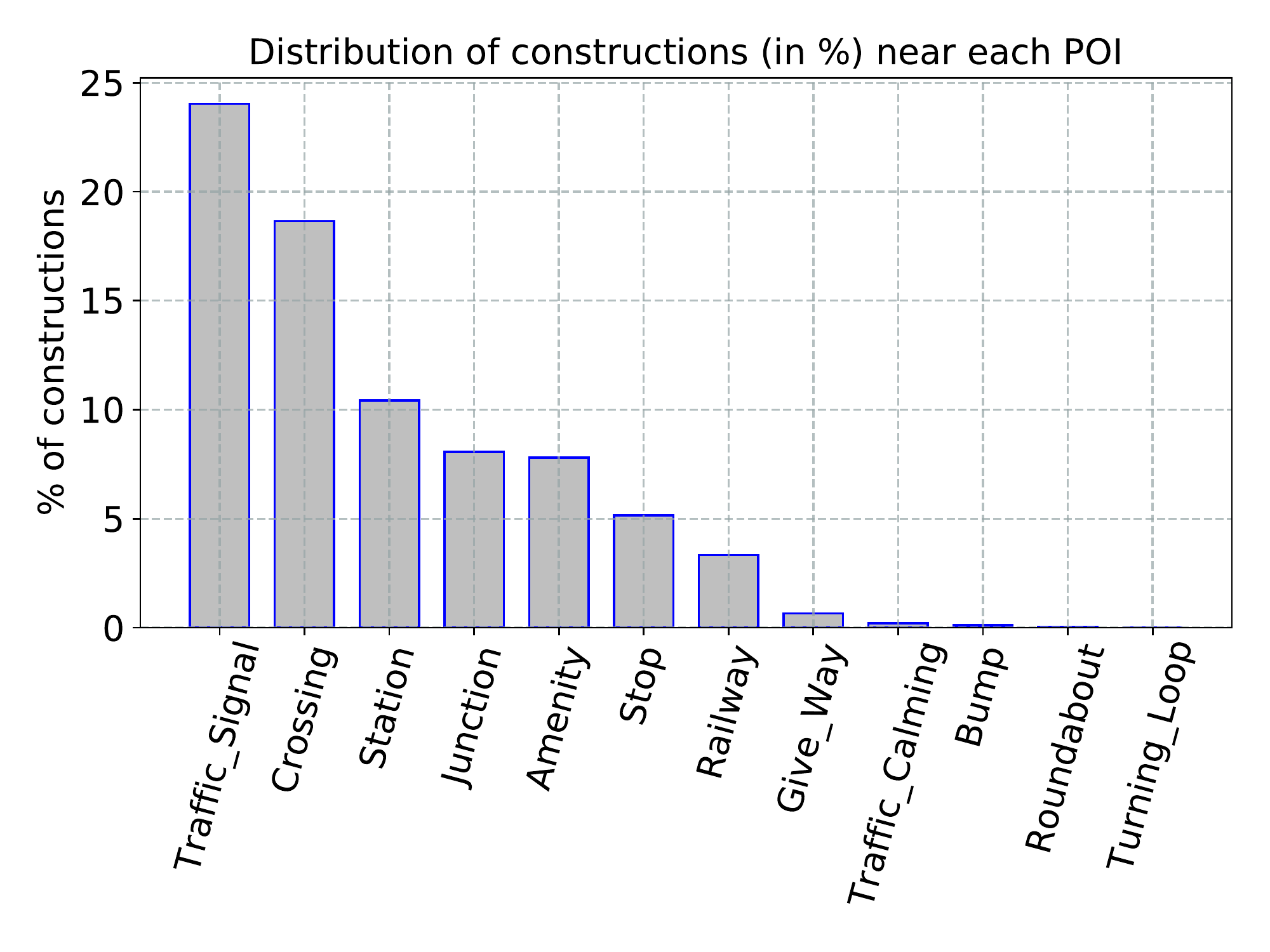}
                \small (d) POI distribution
            \endminipage\hspace{5pt}
            \minipage{0.32\textwidth}
                \centering
                \includegraphics[width=\linewidth]{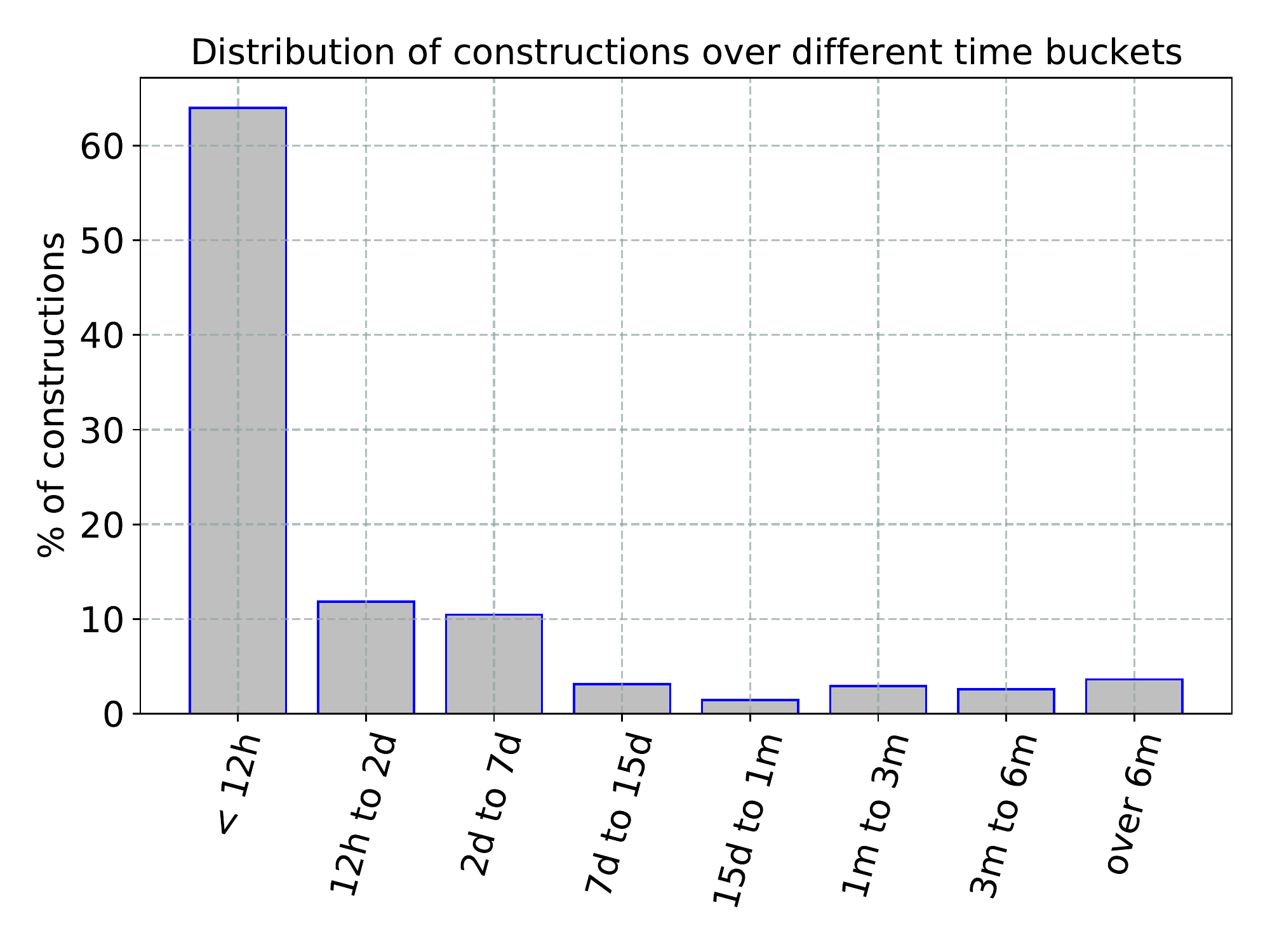}
                \small (e) Duration distribution
            \endminipage\hspace{5pt}
             \minipage{0.32\textwidth}
                \centering
                \includegraphics[width=\linewidth]{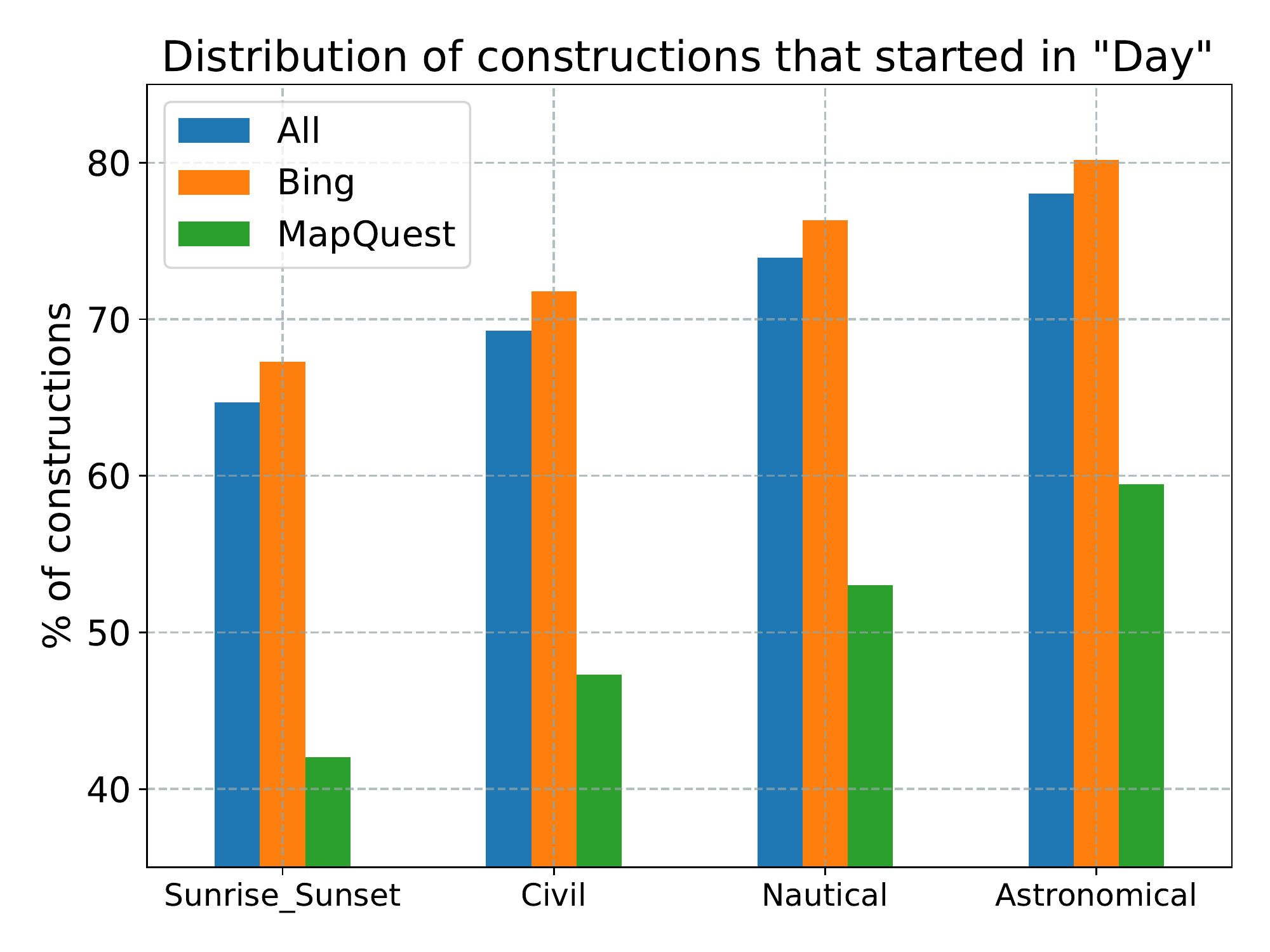}
                \small (f) Daylight distribution
            \endminipage\hfill
            \vspace{-5pt}
            \caption{\small (a) Monthly distribution of constructions based on their start and end time; (b) Distribution of constructions resulting in road closure; (c) OSM-based road-type distribution; (d) OSM-based road annotation type (aka POI) distribution; (e) Duration distribution based on start and end time; and (f) Daylight distribution to show percentage of constructions that started in a day based on four different daylight systems.}
            \label{fig:other_analysis}
        \end{figure*}
    
    \item \textbf{Closure distribution}: According to Figure~\ref{fig:other_analysis}(b), about 15\% of constructions result in a lane closure, while about 5\% resulted in a complete road closure. Also we see constructions reported by MapQuest had more closures, although the overall numbers were dominated by the lower percentages reported by BingMaps. 
    
    \item \textbf{Road-class distribution}: Figure~\ref{fig:other_analysis}(c) shows that high-speed roads such as ``motorway'' and ``primary'' led in construction numbers; however ``residential'' roads (i.e., a low-speed road type) came third. This was an interesting observation, in that it showed that construction takes place on most types of roads. 
        
    \item \textbf{Map annotation (or POI) analysis}: The distribution of constructions annotated by POI in Figure~\ref{fig:other_analysis}(d), showed that a good portion of constructions are reported near intersections. Locations near amenities and stations also make up a big share of the work. 
    
    \item \textbf{Duration and daylight analysis}: As shown in Figure~\ref{fig:other_analysis}(e), over 60\% of constructions only last a few hours. If we consider any construction that lasts 15 days or more to be a long-term construction, then a little over 10\% of constructions are long-term. Further, Figure~\ref{fig:other_analysis}(f) shows the majority of constructions started in the ``day'' based on all four twilight systems. However, constructions reported by MapQuest mostly started during the ``night''. In other words, a good mix of construction times were present in our dataset. 
    
\end{itemize}

\section{Research Question}
\label{sec:problem}
We define our research question in this section. Suppose we are given a set $\mathrm{C}$ of construction events as follows: 
\begin{definition}[Construction event]
    We define a construction event $c$ as $\langle lat, lng, start\_time, end\_time, desc, temp, humid, w\_cond,\\ severity, road\_info, road\_type\rangle$. Here $lat$ and $lng$ are GPS coordinates, $start\_time$ is the time of occurrence, $end\_time$ is the completion time, $desc$ is a human-provided description, $temp$ is the temperature in Fahrenheit, $humid$ is the percentage humidity, $w\_cond$ is the weather condition (e.g., rain, snow, and clear) with a $severity$ value an integer between 1 and 4 (note that all weather features are for the start of the construction), $road\_info$ are additional details about the road (e.g., distance and average speed), and $road\_type$ is the type of the road (e.g., motorway, primary, and secondary). 
\end{definition}

In addition to the above set of construction events, we have a database of geographical map images $\mathrm{M}$ consisting of hexagonal tiles (in any possible view, e.g., satellite, transit, and terrain) with sufficient resolution. Additionally, we have a dataset of points-of-interest $\mathrm{P}$ (e.g., amenities, traffic lights, and stop signs) for a specific zone (or region) of the map. Given these datasets, we define our research question below. 

\vspace{3pt}
\noindent\textbf{Given:}
\begin{itemize}
    \item [--] A set of spatial regions $\mathrm{R} = \{r_1, r_2, \dots, r_n\}$, where $r \in \mathrm{R}$ is a \textit{hexagonal zone}, according to the definition provided by Uber h3\footnote{see \url{https://github.com/uber/h3}} library \citeN{sahr2003geodesic}. Here we choose a resolution level $7$ which results in zones with edge size $1.2 km$ and area $5.16 km^2$. 
    \item [--] A set of fixed-length time intervals $\mathrm{T} = \{t_1, t_2, \dots, t_m\}$, where we set $|t| = 15$ {\it days}, for $t \in \mathrm{T}$.
    \item [--] A database of construction events $\mathrm{C}_r = \{c_1, c_2, \dots\}$ for $r \in \mathrm{R}$. 
    \item [--] A database of map image data $\mathrm{M}_r = \{m_1, m_2, \dots\}$ for $r \in \mathrm{R}$. 
    \item [--] A database of points-of-interest $\mathrm{P}_r = \{p_1, p_2, \dots\}$ for $r \in \mathrm{R}$. 
\end{itemize}

\noindent\textbf{Create:}\vspace{-5pt}
\begin{itemize}
    \item [--] A representation $F_{rt}$ for a region $r \in \mathrm{R}$ during a time interval $t \in \mathrm{T}$, using $C_r$, $M_r$, and $P_r$.
    \item [--] A binary label $L_{rt}$ for $F_{rt}$, where 1 indicates at least one traffic construction was reported during $t$ in region $r$; 0 otherwise. 
\end{itemize}

\noindent\textbf{Find:}\vspace{-5pt}
\begin{itemize}
    \item [--] A model $\mathcal{M}$ to predict $L_{rt}$ using $\langle F_{rt_{i-10}}, F_{rt_{i-9}}, \dots, F_{rt_{i-1}} \rangle$, which means predicting the label of current time interval using observations from the last 10 time intervals.
\end{itemize}

\noindent\textbf{Objective:}\vspace{-5pt}
\begin{itemize}
    \item [--] Minimize the prediction error.  
\end{itemize}

Note that we chose the size of regions and time intervals in order to address the sparsity of input data, while still building a viable model that could provide real-time insights. Interested researchers are encouraged to use different settings to further explore the data and the task. 

\section{Model} 
\label{sec:Construction-Prediction-Model}
\label{sec:method}

This section describes our construction prediction model. We start with a description of the input to the model, that is the feature vector representation of construction events and map image data. Then we describe the model. 

\subsection{Feature Vector Representation} \label{sec:feature-vector-representation}
We create a feature vector representation for each hexagonal geographical region $r$ of resolution $7$ during a time interval $t = 15 \textit{ days}$ by aggregating all the events and averaging over them. To be precise, a construction event includes the following features:

\begin{itemize}
    \item \textbf{Weather (14)}: A vector representing two weather attributes temperature and humidity; and $12$ indicators to represent special weather events light\_rain, moderate\_rain, heavy\_rain, light\_snow, moderate\_snow, heavy\_snow, severe\_cold, severe\_storm, severe\_fog, moderate\_fog, hail, and precipitation\_other. Weather data is obtained from \citeN{moosavi2019short}.
    
    \item \textbf{POI (15)}: A vector of size $15$ to represent frequency of POIs (or map annotations) within $r$. In addition to the cases described in Table~\ref{tab:poi_types}, we also consider \textit{entrance}, \textit{give\_way}, and \textit{turning\_circle}. We obtain POI data from \citeN{osm}.  
    
    \item \textbf{Road type (25)}: A one-hot vector of size $25$ to show the type of the road in region $r$ with construction. Road-type information is also extracted from \cite{osm}. 
    
    \item \textbf{Road information (5)}: On a road segment with a reported construction, this category offers five attributes, namely road segment distance, average speed, approximate travel time, an indicate to show whether the traffic on road segment was impacted during the construction, and severity of the construction. The latter is an integer value between 1 and 4, where 1 indicates the least impact on traffic (i.e., short delay as a result of the event) and 4 indicates a significant impact on traffic (i.e., long delay).
\end{itemize}

To build the aggregated view for all the construction events that occurred during $t$ within region $r$, we simply average over the $59$ attributes described above. Note that except for POIs, other features  are different for the different constructions that we are aggregating over the region. For instance, two constructions that started at different times could have different weather attributes; or they could be associated with different road types. Thus, out of the 59 features, only 15 are constant over time. 

\subsection{Map Image Representation }
\begin{figure}[ht!]
    \centering
    \includegraphics[scale=0.23]{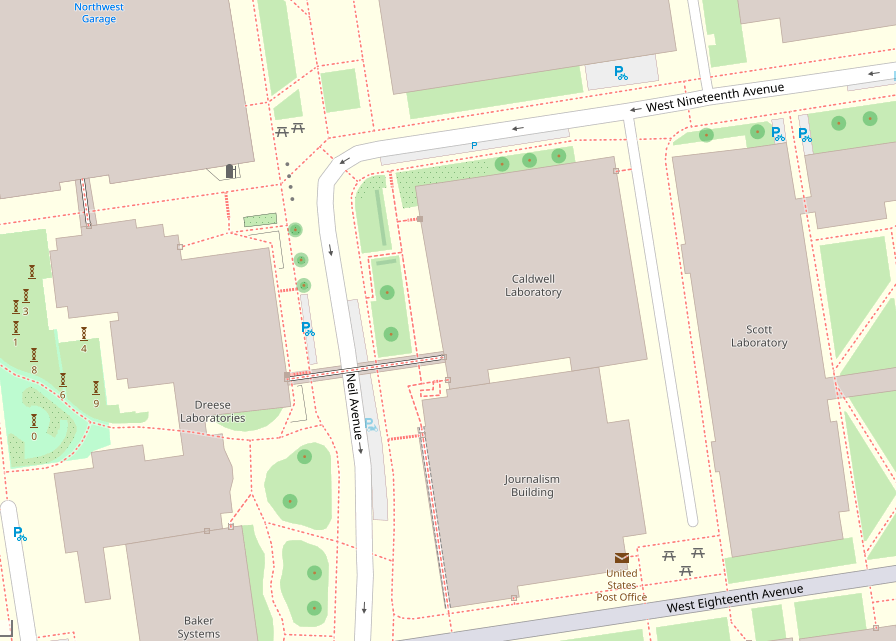}
    \caption{\small Example of a map image extracted from OSM to (roughly) represent a hexagon zone (\textit{zoom\_level = 14})}. 
    \label{fig:osm_map_sample}
\end{figure}

The road network represented in the map tiles is also a relevant context within a spatial encoding of the context for constructions. Constructions could be less prevalent on a road located in a remote area (with a sparse road network), and more prevalent on a road in an urban area (with a dense road network). Figure~\ref{fig:osm_map_sample} shows an example of the type of map images we use, again collected from OpenStreetMap \cite{osm}. We associate each zone with one image. To do so, we first obtain the center of our hexagon in terms of GPS coordinates, and then collect a map tile with the same center from OSM with a zoom level of $14$. According to OSM\footnote{see \url{https://wiki.openstreetmap.org/wiki/Zoom_levels}}, a map tile of \textit{zoom\_level = $14$} is a square image of size $256*256$ pixels. Each pixel of this image covers about $9.547 \textit{ meters}$, that means the side size of the image is about $2.44 \textit{km}$ and its area is about $5.95 \textit{km}^2$. Thus, one single image can almost cover an entire zone, since each side of a zone is about $1.2 \textit{km}$ and its area is about $5.16 \textit{km}^2$. 

\subsection{The Deep Road Construction Prediction (DRCP) Model}
\begin{figure*}
\begin{multicols}{1}
    \begin{subfigure}{1.0\textwidth}
        \includegraphics[width=\linewidth]{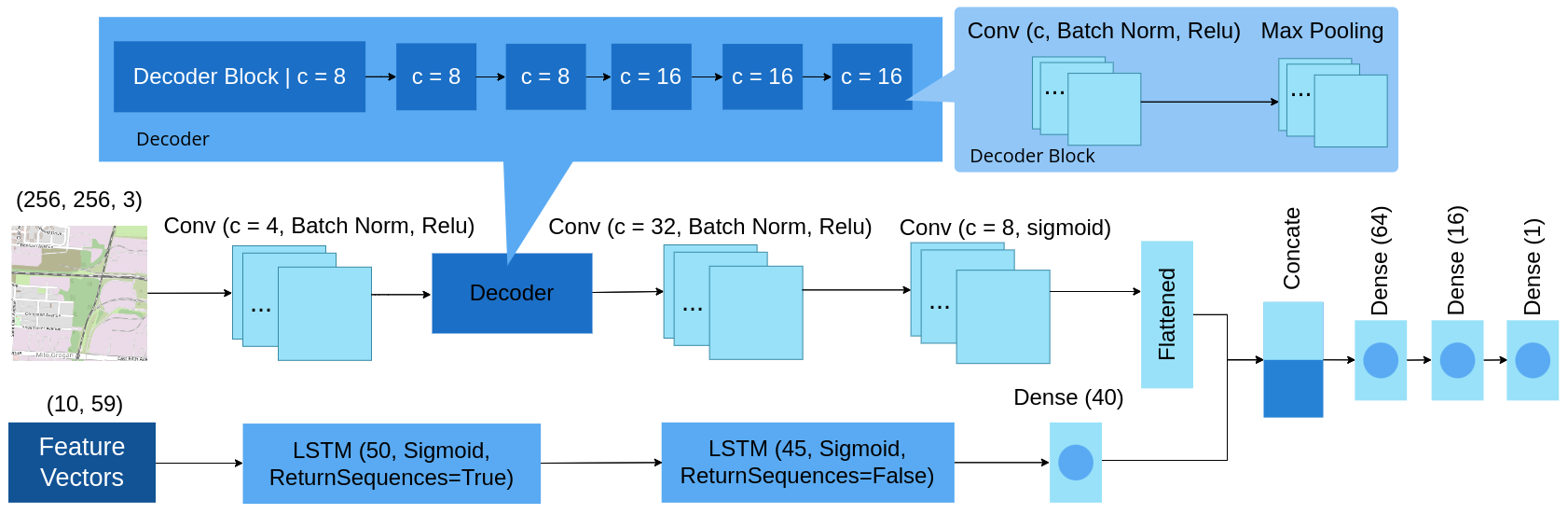}
    \end{subfigure}
\end{multicols}
\caption{Deep Road Construction Prediction (DRCP) Model}
\label{fig:DRCP_model}
\end{figure*}

Our Deep Road Construction Prediction (DRCP) model is shown in Figure \ref{fig:DRCP_model}.
Since our input data contains both images and time-series data we use a mix of convolutional (CNN) \citeN{wu2017introduction}, recurrent (RNN) \citeN{lei2021rnn} and fully connected components. 

\begin{itemize}[leftmargin=0.3cm]
    \item \textbf{CNN component}: The use of this component is to encode map image data to extract latent spatial features. The size of an input image is $256\times256\times3$. The $CNN$ component built from these images is shown in Figure~\ref{fig:DRCP_model}, and it comprises four sub-components, three convolutional blocks (with 4, 32, and 8 channels, respectively) and one \textit{decoder} component. The decoder component contains six decoder blocks, which are convolutional layers with $8$ channels in the first and $16$ channels in the last three blocks. All decoder blocks include batch normalization \citeN{santurkar2018does} to deal with internal covariate shift, max pooling \citeN{graham2014fractional} for downsampling, and \textit{ReLU} \citeN{agarap2018deep} as an activation function. The other sub-components (i.e., three convolutional blocks) do not leverage max-pooling, but batch normalization is used in two of them. The output of the CNN component is then converted to a vector of size $128$ by using a flatten layer. Note that the activation function used in the last sub-component is \textit{sigmoid} to properly concatenate the outputs of CNN and RNN components. The kernel size in all convolutional layers is $3\times3$, and stride size is $1$. It is worth noting that this design is mostly driven empirically through exploring different architectures and settings. 
    
    \item \textbf{RNN component}: To encode sequential data, we use two layers of Long Short Term Memory $(LSTM)$ \citeN{zhao2017lstm} with $59$ and $45$ neurons, respectively. Both layers of $LSTM$ use \textit{sigmoid} as activation function. The choice of $59$ neurons in the first LSTM layer is to utilize sequential input of size $10 \times 59$ that represents aggregated construction event data over the past $10$ time intervals. The output of this component is then converted to a vector of size $40$ by using a dense layer. We employed \textit{sigmoid} activation function for the dense layer too. 
    
    \item \textbf{Fully connected component}: This component includes three layers of size $64$, $16$, and $1$. The input to this layer is simply a concatenation of the outputs of the other two components, which is a vector of size $168$. The final output of the fully connected component shows whether there will be a construction in the next $15$ days (i.e., next time interval) or not. 
\end{itemize}

We employed an extensive grid-search-based hyper-parameter tuning process to find the best settings (e.g., layer size and network size) for each component. 

\section{Experiment and Results}
\label{sec:results}
In this section we first describe experiment setup, then describe data, followed by baseline models description, and conclude it by presenting results and discussions\footnote{All codes and sample data are available at \url{https://github.com/7Amin/DRCP}}. 

\subsection{Experiment Setup}
All implementations\footnote{see our code on  \url{https://github.com/7Amin/DRCP}} are in Python using Tensorflow \citeN{abadi2016tensorflow}, Keras \citeN{chollet2015keras}, and scikit-learn \citeN{JMLR:v12:pedregosa11a} libraries; and experiments were run on machines at Ohio Supercomputer Center \citeN{OhioSupercomputerCenter1987}. 
For training \textit{DRCP} we used the \textit{Adam} \citeN{kingma2014adam} optimizer. The maximum number of epochs was set to be $1000$, with an early stopping policy where training stops if the loss value on the validation set does not decrease after $30$ consecutive epochs. The initial learning rate was set to $10^{-4}$, and if no improvements could be observed for $5$ consecutive epochs, then the learning rate is reduced by a factor of $0.9$ (i.e., $\textit{new learning rate } = \textit{learning rate} \times 0.9$). This reduction could potentially continue until the learning rate reaches $10^{-6}$. 
The loss function we used was \textit{Binary Cross Entropy} \citeN{ketkar2017introduction}, since our problem is a binary classification problem and this loss function is proven to work quite well for this class of problems.

\subsection{Data Description}
We trained and validated our \textit{DRCP} model, on eight states (\textit{New York}, \textit{Pennsylvania}, \textit{Georgia}, \textit{Texas}, \textit{Colorado}, \textit{Florida}, \textit{Washington} and \textit{Michigan}). Further, we selected nine cities (\textit{Columbus (OH)}, \textit{New York City (NY)}, \textit{Pittsburgh (PA)}, \textit{Atlanta (GA)}, \textit{Houston (TX)}, \textit{Denver (CO)}, \textit{Miami (FL)}, \textit{Seattle (WA)}, and \textit{Detroit (MI)}. The choice of these states and cities was primarily to achieve diversity in traffic and weather conditions, population, population density, and urban characteristics (road-network, prevalence of urban versus highway roads, etc.). We split our data set into three parts. The first part was our training set with records drawn from  the date range of February 2016 to the end of December 2019. The second part was our validation set, whose dates ranged from January 2020 to the end of May 2020. The last part of our data ranged from June 2020 to the end of December 2020 and was used as our as test set.

We prepared our data for use in the machine learning process as described in Section~\ref{sec:method}. Each record of data includes aggregated feature vector representation for a $15$ days time interval, as well as a map image to represent the corresponding geographical zone. Note that our goal is to predict a binary label for the next interval, using data from the past $10$ intervals. To mitigate label imbalance, we use \textit{class weights} to better train different models. We empirically found the weights $1.01$ and $16.01$ for the classes $0$ and $1$, respectively. These weights were found from the use of the validation data in the DRCP model. 

\subsection{Baseline Models}
We choose models described below as baselines to compare our proposal against them. 
\begin{itemize}[leftmargin=0.3cm]
    \item \textbf{Logistic Regression (LR)}: LR is a well-known model for binary classification tasks. To train the model we set \textit{penalty = l2}, \textit{random state = 0}, and used ``Limited-memory Broyden Fletcher Goldfarb Shanno (lbfgs)'' as the solver. 
    
    \item \textbf{Random Forest(RF)}: Given the nature of our input data, RF seems a natural choice to be used for the task defined in this paper. To train this model, we set \textit{number of estimators = 10}. 
    
    \item \textbf{Gradient Boosting Classifier (GB)}: GB is a strong tree-based boosting model for a wide range of classification tasks, thus we choose it as a baseline in this work. We set \textit{number of estimators = 10} to train the model. 
    
    \item \textbf{Multi-layer Perceptron (MLP)}: MLP is another strong choice to compare our proposal against for the construction prediction task. Here we use a hidden layer with $100$ neurons, \textit{Relu} as activation function, and \textit{Adam} as optimizer to train the model. 
    
\end{itemize}

Since these baseline models cannot work with sequential data, we vectorize inputs and feed them as a single vector to these models. The vectorization process makes use of both construction events data as well as map image data in a concatenated form.

\subsection{Results and Discussions}
In this section we define two experimentation scenarios and present results for each. The first scenario considers the case where construction data is potentially available for every zone in a region, thus training and inference can be done based on input data from all zones. The second scenario, on the other hand, assumes we only have data for some of zones in a region, but still seek to make future predictions for all zones (i.e., sparsity). 

\subsubsection{Scenario I: training based on all zones}
\label{subsec:scn1}
According to this scenario, the input data to train different models includes all the zones that have any data to offer. Using the settings described in previous sections, we trained all the models, and Tables~\ref{tab:normal_result_city_F-score} and \ref{tab:normal_result_city_Accuracy} show the results based on \textit{f1-score} and \textit{accuracy} metrics, and for the selected cities. Our proposed model (i.e., DRCP) outperforms other models with significant margin for 8 out of 9 cities based on both f1-score and accuracy metric. On average, we observe $3.2\%$ improvement in accuracy and $2.8\%$ improvement in F1-score when compared to the best baseline for each city. This demonstrates the effectiveness of our proposal to make better use of spatiotemporal information to make future predictions. 

\begin{table}[h!]
    \small 
    \centering
    \setlength\tabcolsep{7pt}
    \caption{Comparing different models for selected cities for Scenario I using ``F1-score''}\vspace{-5pt}
    \begin{tabular}{|c|c|c|c|c|c|}
    \hline
   \rowcolor{Gray} \textbf{City} &  \textbf{LR} & \textbf{RF} & \textbf{GB} & \textbf{MLP} & \textbf{DRCP} \\
    \hline
        Atlanta (GA)  &  $0.773$ & $0.751$ & $0.731$ & $0.725$ &   $\textbf{0.793}$\\
    \hline
        Columbus (OH)  &  $0.859$ & $0.87$ & $0.855$ & $0.891$ &   $\textbf{0.902}$ \\
    \hline
        Denver (CO) & $0.756$ & $0.767$ & $0.753$ & $0.761$ &   $\textbf{0.794}$ \\
    \hline
        Detroit (MI) & $0.774$ & $0.786$ & $0.743$ & $0.728$ &  $\textbf{0.795}$\\
    \hline
        Houston (TX)  & $0.891$ & $0.895$ & $0.892$ & $0.884$ &   $\textbf{0.908}$\\
    \hline
        Miami (FL) & $0.856$ & $0.843$ & $0.845$ & $\textbf{0.868}$  &  $0.845$ \\
     \hline
        New York City (NY)  &  $0.895$ & $0.898$ & $0.884$ & $0.906$ &   $\textbf{0.942}$ \\
    \hline
        Pittsburgh (PA)  &  $0.798$ & $0.794$ & $0.782$ & $0.768$ &   $\textbf{0.832}$\\
    \hline
        Seattle (WA)  & $0.852$ & $0.825$ & $0.803$ & $0.794$ &  $\textbf{0.853}$\\
    \hline
    \end{tabular}
    \label{tab:normal_result_city_F-score}
\end{table}

\begin{table}[h!]
    \small
    \centering
    \setlength\tabcolsep{7pt}
    \caption{Comparing different models based on Scenario I using ``Accuracy'' and for the selected cities}\vspace{-5pt}
    \begin{tabular}{|c|c|c|c|c|c|}
    \hline
   \rowcolor{Gray} \textbf{City} &  \textbf{LR} & \textbf{RF} & \textbf{GB} & \textbf{MLP} & \textbf{DRCP} \\
    \hline
        Atlanta (GA)  &  $74.85\%$ & $72.55\%$ & $70.32\%$ & $69.44\%$  &   $\textbf{77.9\%}$\\
    \hline
        Columbus (OH)  & $82.3\%$ & $83.19\%$ & $82.81\%$ & $86.11\%$  &   $\textbf{87.2\%}$\\
    \hline
        Denver (CO) & $73.56\%$ & $74.53\%$ & $73.48\%$ & $73.68\%$  &   $\textbf{78.2\%}$\\
     \hline
        Detroit (MI)  & $74.46\%$ & $75.36\%$ & $70.9\%$ & $68.47\%$ &   $\textbf{78.2\%}$\\
    \hline
        Houston (TX) &  $86.68\%$ & $86.8\%$ & $86.9\%$ & $86.21\%$ &   $\textbf{88.1\%}$ \\
    \hline
        Miami (FL) & $82.86\%$ & $81.09\%$ & $82.17\%$ & $\textbf{83.91\%}$ &   $81.6\%$ \\
    \hline
        New York City (NY)  &  $85.01\%$ & $83.95\%$ & $85.41\%$ & $86.33\%$ &   $\textbf{89.7\%}$\\
    \hline
        Pittsburgh (PA) &  $76.18\%$ & $74.82\%$ & $\%75.53$ & $71.9\%$  &   $\textbf{77.7\%}$\\
    \hline
        Seattle (WA)  & $81.39\%$ & $79.23\%$ & $77.52\%$ & $75.81\%$ &   $\textbf{81.43\%}$\\
    \hline
    \end{tabular}
    \label{tab:normal_result_city_Accuracy}
\end{table}

\begin{figure*}
    \begin{multicols}{3}
        \begin{subfigure}{0.26\textwidth}
            \includegraphics[width=\linewidth]{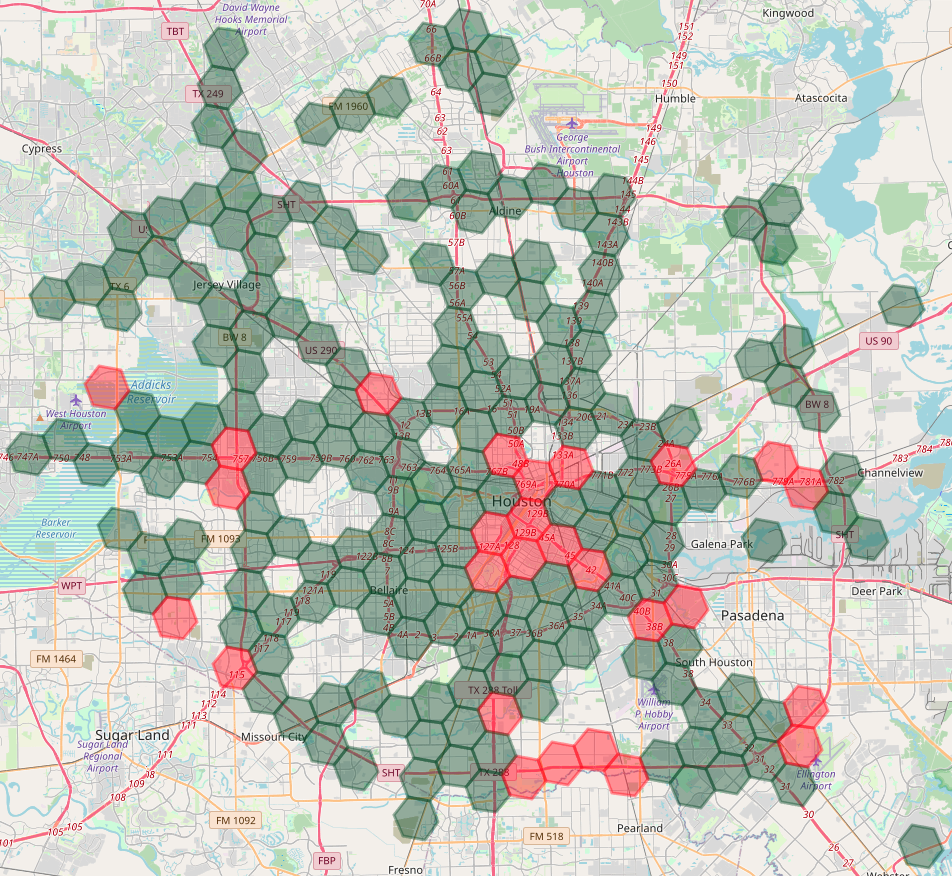}
            \caption{Predicted / 6-1-2020 to 6-15-2020}
        \end{subfigure}\par 
        \begin{subfigure}{0.26\textwidth}
            \includegraphics[width=\linewidth]{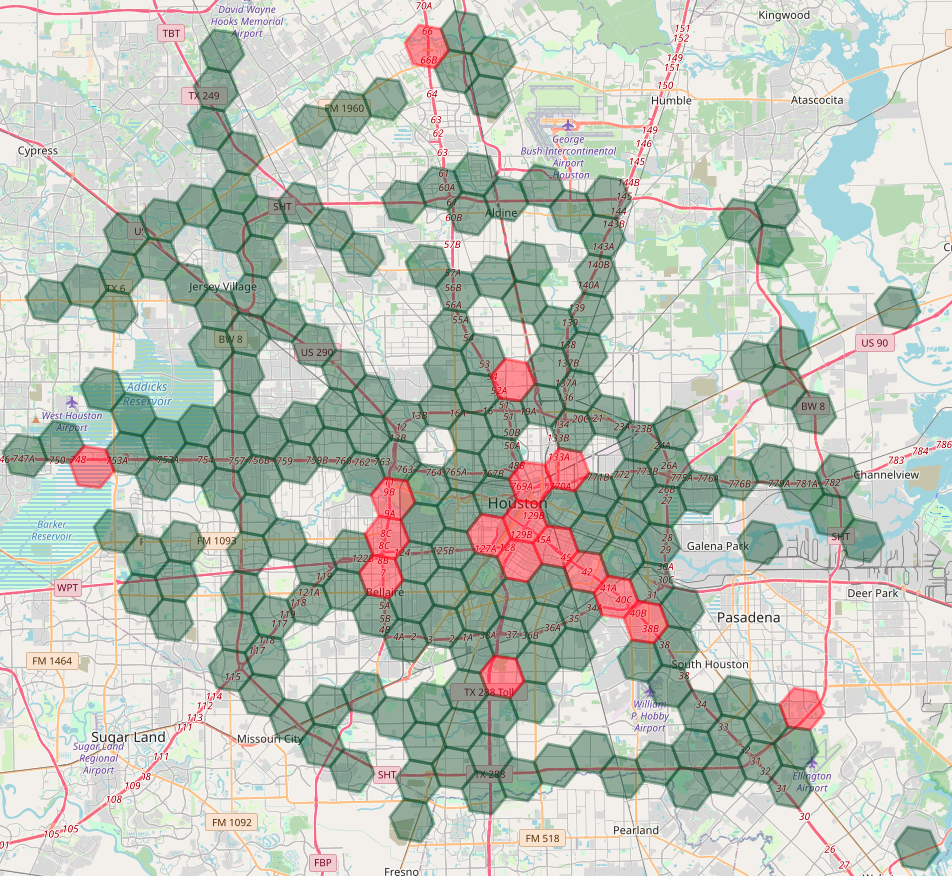}
            \caption{Predicted / 6-16-2020 to 6-30-2020}
        \end{subfigure}\par 
        \begin{subfigure}{0.26\textwidth}
            \includegraphics[width=\linewidth]{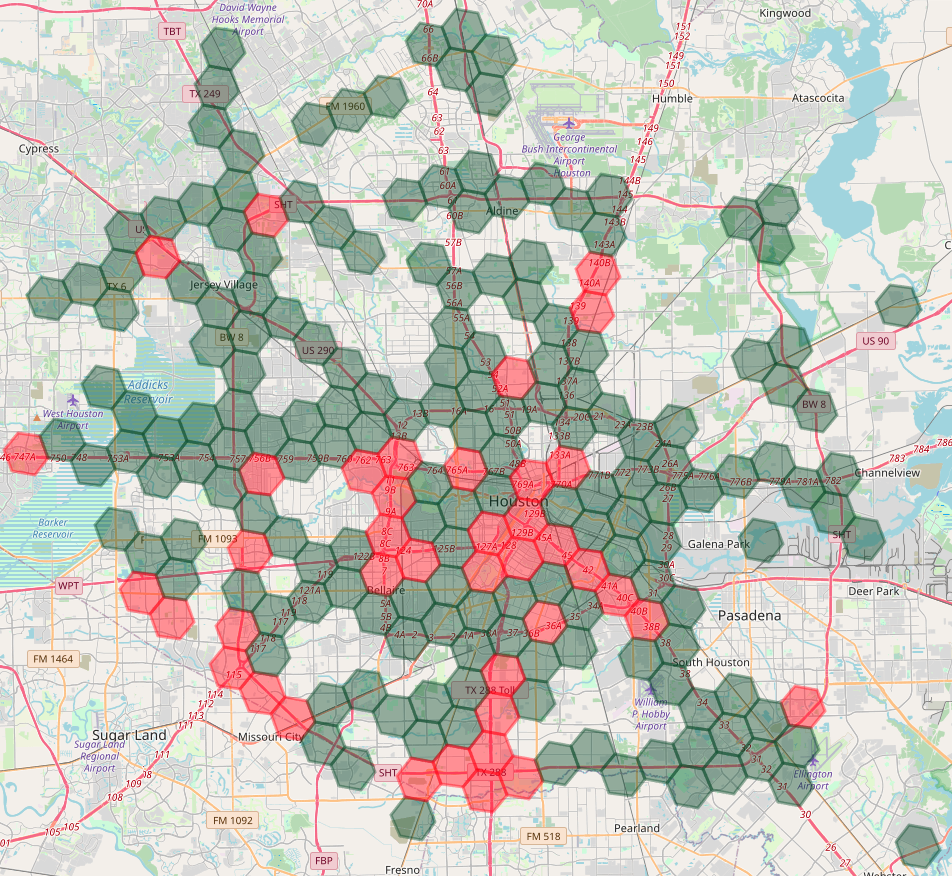}
            \caption{Predicted / 7-1-2020 to 7-15-2020}
        \end{subfigure}\par 
    \end{multicols}
    \begin{multicols}{3}
        \begin{subfigure}{0.26\textwidth}
            \includegraphics[width=\linewidth]{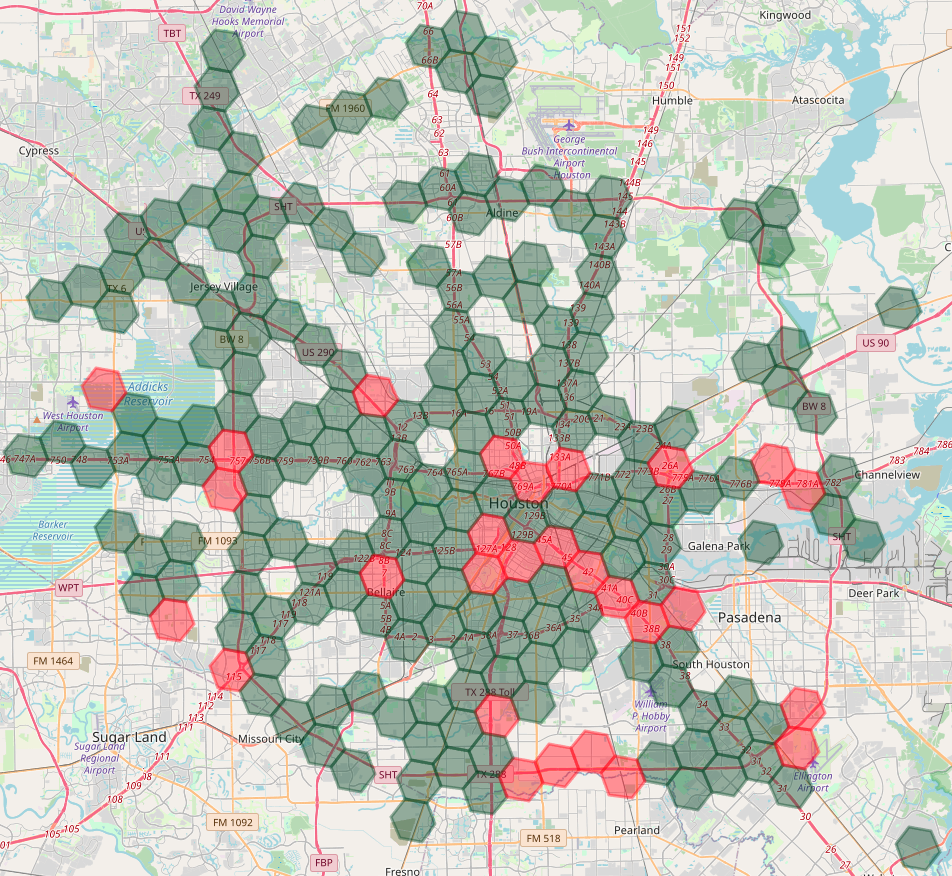}
            \caption{Actual / 6-1-2020 to 6-15-2020}
        \end{subfigure}\par 
        \begin{subfigure}{0.26\textwidth}
            \includegraphics[width=\linewidth]{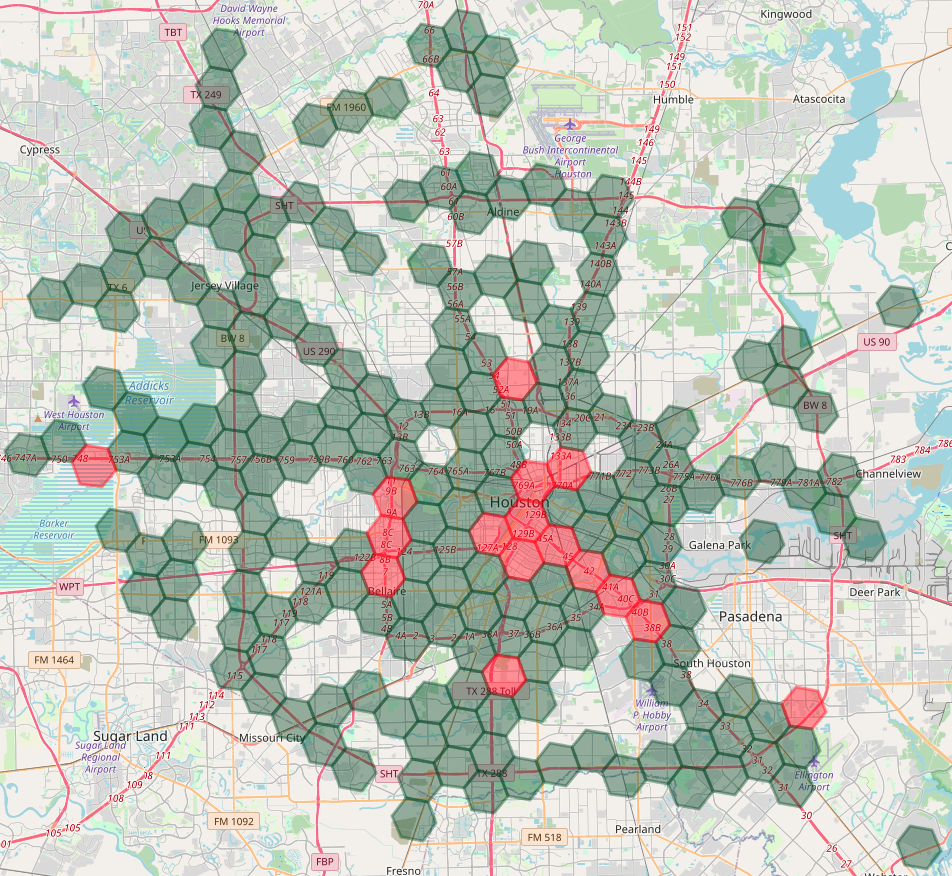}
            \caption{Actual / 6-16-2020 to 6-30-2020}
        \end{subfigure}\par 
        \begin{subfigure}{0.26\textwidth}
            \includegraphics[width=\linewidth]{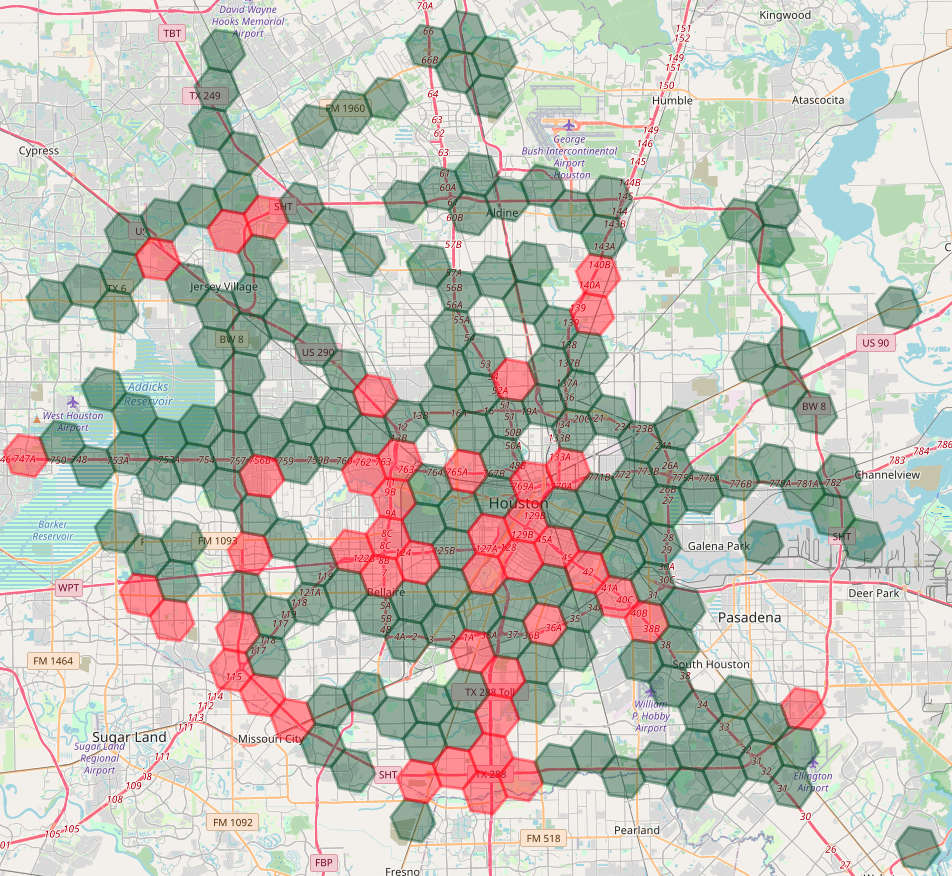}
            \caption{Actual / 7-1-2020 to 7-15-2020}
        \end{subfigure}\par 
    \end{multicols}
    \vspace{-5pt}
    \caption{\small Example of prediction results (a to c) by the ``DRCP'' model along with actual labels (d to f) over three consecutive time frames for Houston in Texas. Here we make predictions for individual zones that are represented by hexagons. A ``green'' hexagon shows no construction is reported or predicted during the corresponding time frame, and a ``red'' hexagon shows otherwise.}
    \label{fig:result_texas}
\end{figure*}

To visualize how our model performs, Figure~\ref{fig:result_texas} shows prediction results (a, b, and c) using the DRCP model and expected outcomes (d, e, and f) for Houston (TX) over three consecutive time frames. A zone represented by ``green'' indicates there are no road constructions reported or predicted, while a zone represented by ``red'' shows otherwise. It is interesting to see how the model performs over different time frames with different distributions of constructions. For instance, Figure~\ref{fig:result_texas}-(e) shows more sparsity, while Figure~\ref{fig:result_texas}-(f) is more dense; and in both cases our proposed model performs reasonably to make future predictions (see Figure~\ref{fig:result_texas}-(b) and Figure~\ref{fig:result_texas}-(c), respectively). This shows how the proper use of useful spatiotemporal information can result in satisfactory model outcomes, which is a strong indicator of applicability of this setup in the real-world. In Section~\ref{sec:appendix} we provided further visualized results of this kind for two other cities in the United States (see Figures~\ref{fig:result_ohio} and \ref{fig:result_florida}). 

\subsubsection{Scenario II: training based on limited zones}
Unlike the first scenario, here the assumption is only a limited number of zones have data to offer for training. To do so, we randomly select $60\%$ of our zones for training, $20\%$ for validation, and $20\%$ for testing. To make the task even more challenging, we do state-level training and testing (as opposed to city-level as in scenario I), which results in more sparsity, and perhaps closer to a real-world setup. Using a similar training process, we individually trained the \textit{DRCP} model for the selected states, and Table~\ref{tab:location_split_states} shows the results based on ``F1-score'' and ``Accuracy'' metrics. Based on the results, we see our model performs quite reasonably when predicting on test data that include held-out zones. This is an important observation since it shows that even a partial dataset collected from a limited number of zones in a state is enough to train a generalizable model like DRCP to make inferences about future constructions for all zones in a state. Additionally, this is useful in real-world, where we may never have access to high quality data for all zones in a state or even in a city. Lastly, we note that similarities in input data (e.g., map structure and weather condition) within a state (at least for the states that we experimented with) could be another reason to justify the promising outcomes of our model in this scenario. 

\begin{table}[ht!]
    \centering
    \setlength\tabcolsep{7pt}
    \caption{Prediction results using DRCP model based on Scenario II (i.e., studying spatial sparsity) for selected states}\vspace{-5pt}
    \begin{tabular}{|c|c|c|}
    \hline
   \rowcolor{Gray} \textbf{State} & \textbf{F1-score}  & \textbf{Accuracy}\\
    \hline
        Colorado &   $0.934$ &  $84.4\%$  \\
    \hline
        Florida &   $0.942$ & $85.7\%$ \\
    \hline
        Georgia &   $0.942$ &  $88.1\%$ \\
    \hline
        Michigan &   $0.941$ & $88.2\%$ \\
    \hline
        New York &   $0.925$ & $84.6\%$ \\
    \hline
        Pennsylvania &   $0.912$ &  $85.0\%$ \\
    \hline
        Texas &   $0.938$ &  $84.7\%$ \\
    \hline
        Washington &   $0.947$ & $88.9\%$  \\
    \hline
    \end{tabular}
    \label{tab:location_split_states}
\end{table}

\section{Conclusion and Future Work}
\label{sec:conlcusion}
In this paper we tackle the problem of future constructions prediction using heterogeneous spatiotemporal information such as past constructions, weather data, and geographical map data. To our knowledge, this is a relatively new problem space that has not been explored by the research community, maybe due to lack of comprehensive historical data about past constructions. To address this gap, this paper introduces a novel dataset of 6.2 million constructions in the United States between 2016 and 2021, that offers a variety of details around location, time, weather condition, map, and road-network. 
Additionally, we formulate and solve the problem of predicting the possibility of future constructions using such data. We present a deep-neural-network-based model to efficiently utilize the heterogeneous input by combining convolutional and recurrent components in a reasonable setting. Through extensive experiments over multiple major cities and states in the United States, we show the usefulness of our proposal in a real-world setting and in comparison to several state-of-the-art baselines. 

As directions for future research, we can extend our input data and leverage information such as traffic load, past traffic accidents, finer-grained weather data, demographic data for each zone, as well as other map imagery views (e.g., satellite view). In addition to data, the model that we presented in this paper can potentially be improved by jointly modeling spatial and temporal data, instead of a separate utilization. Lastly, we can extend the task that is defined in this work and tackle more advanced problems such as ``construction type'' prediction; that is, if a certain construction could result in a closure or not.

\bibliographystyle{ACM-Reference-Format}
\bibliography{references}

\pagebreak
\onecolumn
\section{Appendix}
\label{sec:appendix}

\subsection{Visualizing road construction predictions}
\begin{figure*}[h]
    \begin{multicols}{3}
        \begin{subfigure}{0.26\textwidth}
            \includegraphics[width=\linewidth]{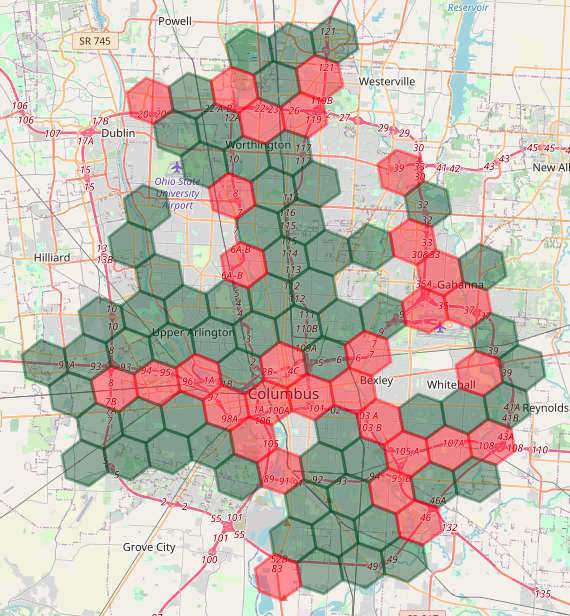}
            \caption{Predicted / 6-1-2020 to 6-15-2020}
        \end{subfigure}\par 
        \begin{subfigure}{0.26\textwidth}
            \includegraphics[width=\linewidth]{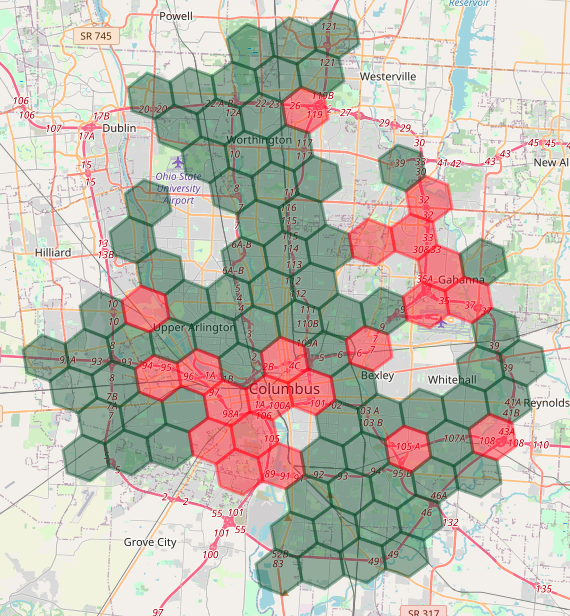}
            \caption{Predicted / 6-16-2020 to 6-30-2020}
        \end{subfigure}\par 
        \begin{subfigure}{0.26\textwidth}
            \includegraphics[width=\linewidth]{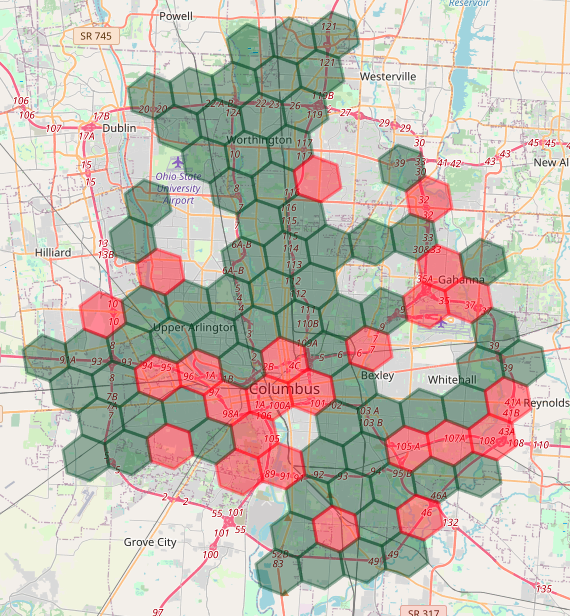}
            \caption{Predicted / 7-1-2020 to 7-15-2020}
        \end{subfigure}\par 
    \end{multicols}
    \begin{multicols}{3}
        \begin{subfigure}{0.26\textwidth}
            \includegraphics[width=\linewidth]{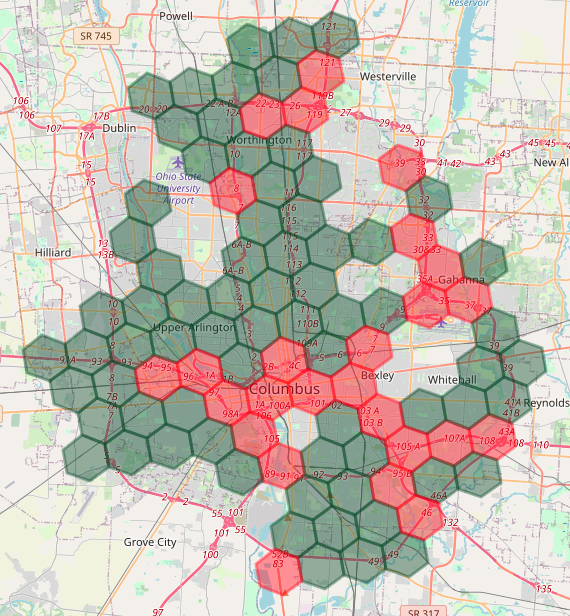}
            \caption{Actual / 6-1-2020 to 6-15-2020}
        \end{subfigure}\par 
        \begin{subfigure}{0.26\textwidth}
            \includegraphics[width=\linewidth]{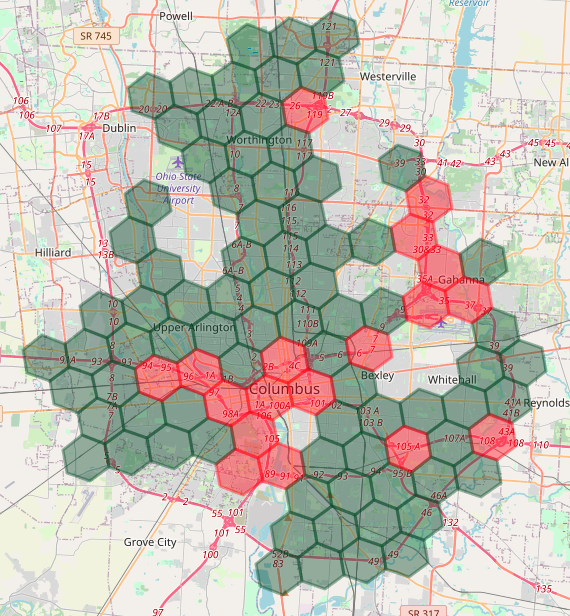}
            \caption{Actual / 6-16-2020 to 6-30-2020}
        \end{subfigure}\par 
        \begin{subfigure}{0.26\textwidth}
            \includegraphics[width=\linewidth]{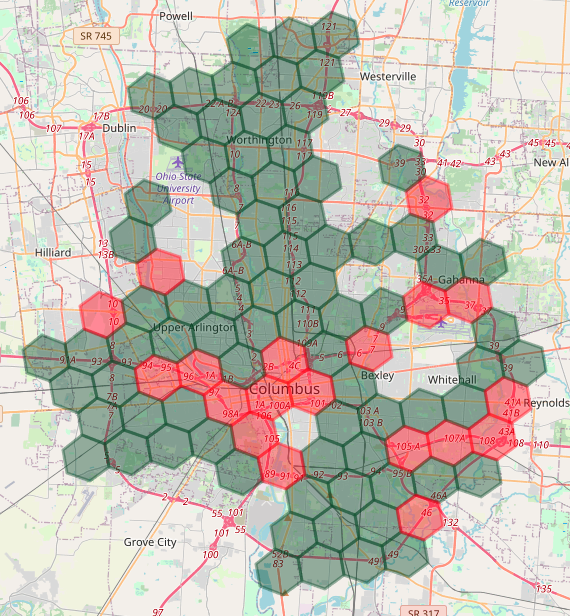}
            \caption{Actual / 7-1-2020 to 7-15-2020}
        \end{subfigure}\par 
    \end{multicols}
    \vspace{-5pt}
    \caption{\small Example of prediction results (a to c) by the ``DRCP'' model along with actual labels (d to f) over three consecutive time frames for Columbus in Ohio. Here we make predictions for individual zones that are represented by hexagons. A ``green'' hexagon shows no construction is reported or predicted during the corresponding time frame, and a ``red'' hexagon shows otherwise.}
    \label{fig:result_ohio}
\end{figure*}

\begin{figure*}
    \begin{multicols}{3}
        \begin{subfigure}{0.26\textwidth}
            \includegraphics[width=\linewidth]{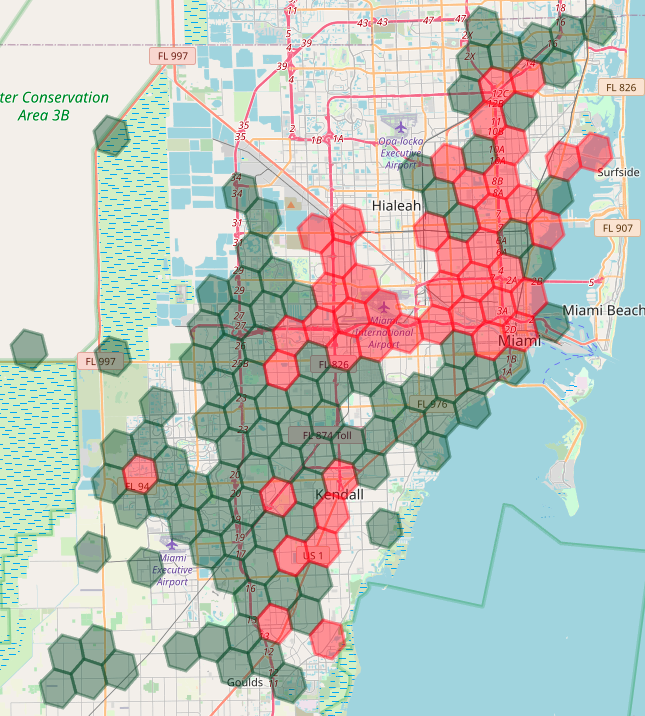}
            \caption{Predicted / 6-1-2020 to 6-15-2020}
        \end{subfigure}\par 
        \begin{subfigure}{0.26\textwidth}
            \includegraphics[width=\linewidth]{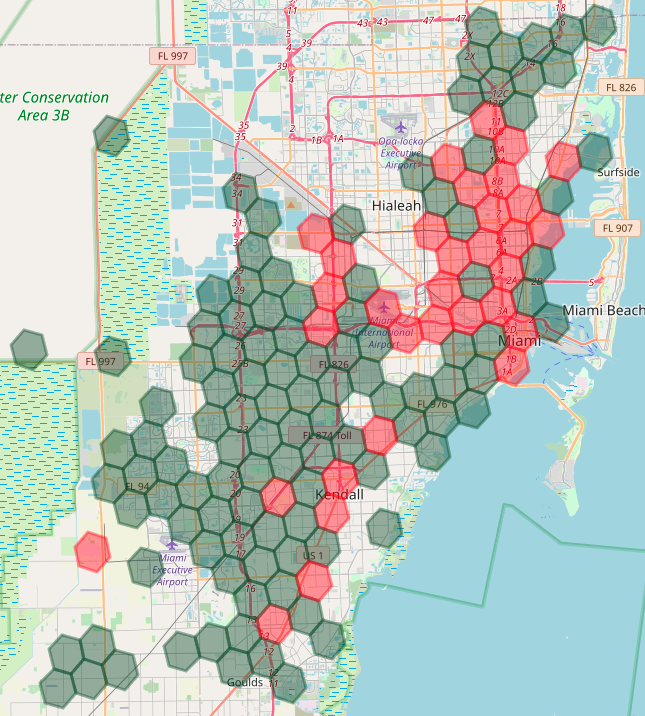}
            \caption{Predicted / 6-16-2020 to 6-30-2020}
        \end{subfigure}\par 
        \begin{subfigure}{0.26\textwidth}
            \includegraphics[width=\linewidth]{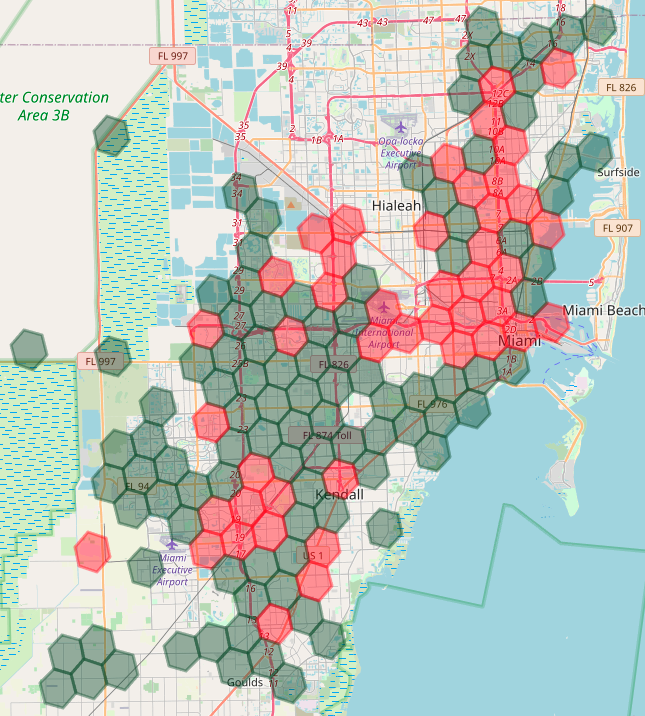}
            \caption{Predicted / 7-1-2020 to 7-15-2020}
        \end{subfigure}\par 
    \end{multicols}
    \begin{multicols}{3}
        \begin{subfigure}{0.26\textwidth}
            \includegraphics[width=\linewidth]{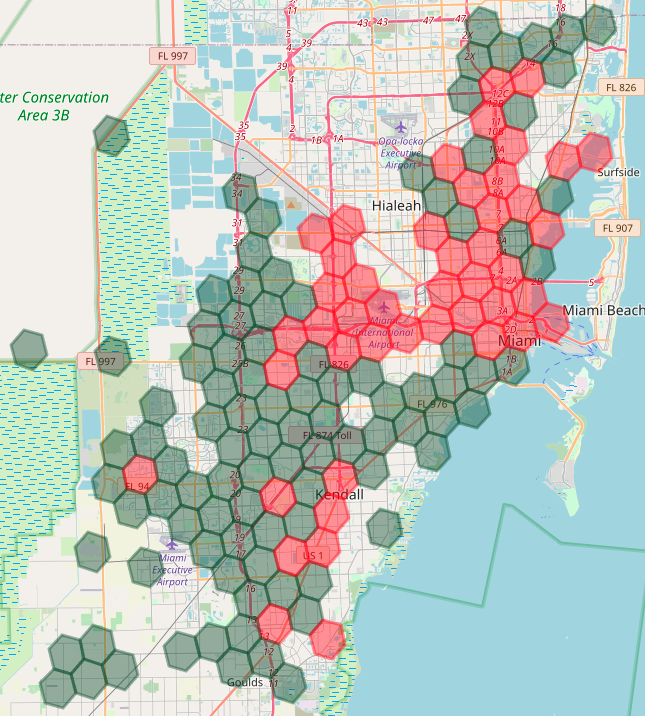}
            \caption{Actual / 6-1-2020 to 6-15-2020}
        \end{subfigure}\par 
        \begin{subfigure}{0.26\textwidth}
            \includegraphics[width=\linewidth]{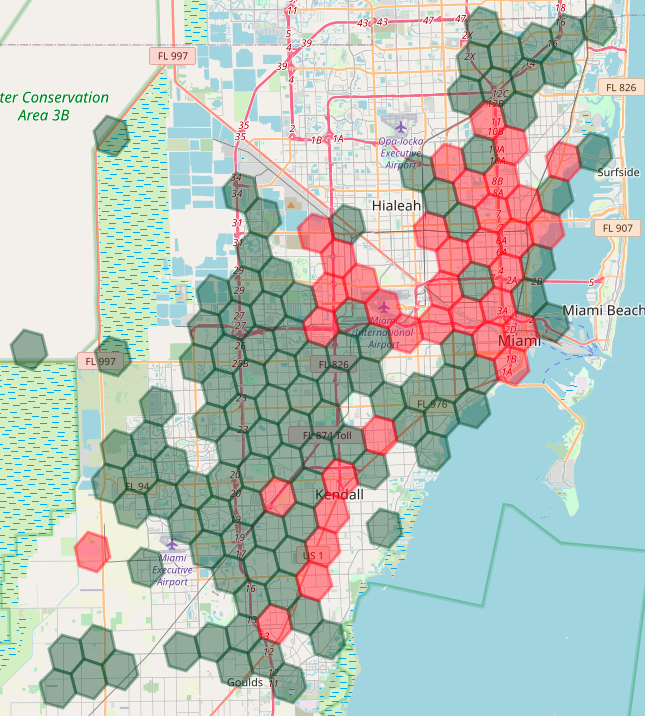}
            \caption{Actual / 6-16-2020 to 6-30-2020}
        \end{subfigure}\par 
        \begin{subfigure}{0.26\textwidth}
            \includegraphics[width=\linewidth]{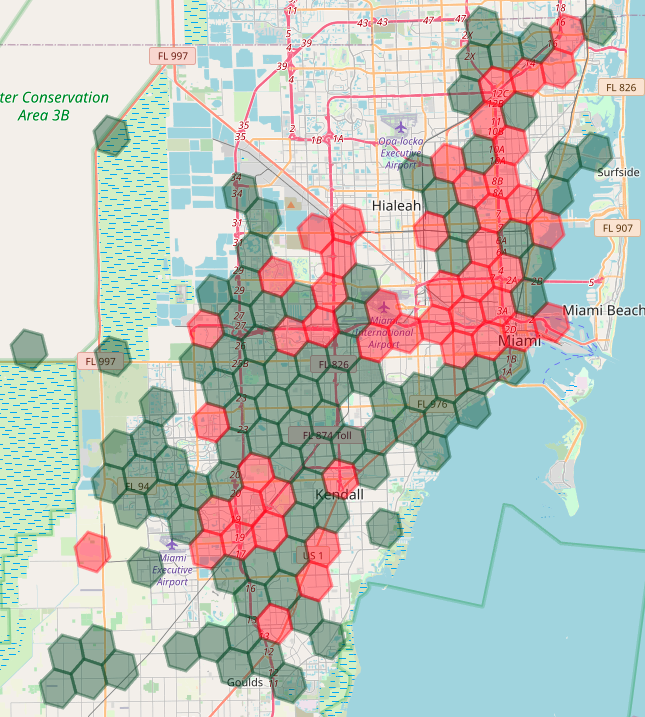}
            \caption{Actual / 7-1-2020 to 7-15-2020}
        \end{subfigure}\par 
    \end{multicols}
    \vspace{-5pt}
    \caption{\small Example of prediction results (a to c) by the ``DRCP'' model along with actual labels (d to f) over three consecutive time frames for Miami in Florida. Here we make predictions for individual zones that are represented by hexagons. A ``green'' hexagon shows no construction is reported or predicted during the corresponding time frame, and a ``red'' hexagon shows otherwise.}
    \label{fig:result_florida}
\end{figure*}

In this section we visualize road construction prediction results for Columbus (in Ohio) and Miami (in Florida) by Figures \ref{fig:result_ohio} and \ref{fig:result_florida}, respectively. Settings and details are the same as what we earlier described in Section~\ref{subsec:scn1}. Similar to what we presented before, we can see how our proposed model performs future construction prediction in two different cities. It is worth noting again that some of the chosen time frames are sparse, given the number of reported constructions during those. However, our proposed model is capable of making solid predictions even when there is sparsity. 

\end{document}